%% file: main.tex
\documentclass[10pt,twocolumn,letterpaper]{article}

\usepackage{iccv}
\usepackage{times}
\usepackage{epsfig}
\usepackage{graphicx}
\usepackage{amsmath}
\usepackage{amssymb}

\usepackage{times}
\usepackage{epsfig}
\usepackage{graphicx}
\usepackage{amsmath}
\usepackage{amssymb}

\usepackage[utf8]{inputenc} %
\usepackage[T1]{fontenc}    %

\usepackage{url}            %
\usepackage{booktabs}       %
\usepackage{amsfonts}       %
\usepackage{nicefrac}       %
\usepackage{microtype}      %
\usepackage{graphicx}
\usepackage{multirow}
\usepackage{subfigure}
\usepackage{amsmath}
\usepackage{floatrow}
\usepackage{caption}

\usepackage{bm}%
\usepackage{url}
\usepackage{color}
\usepackage{mathrsfs}
\usepackage{amsthm}
\usepackage{float}
\floatstyle{plaintop}
\restylefloat{table}

\usepackage{mathtools}
\usepackage{enumitem}
\usepackage{romannum}
\usepackage{combelow}
\usepackage{textcomp}

\usepackage{cases}

\newtheorem*{theorem*}{Theorem}

\usepackage{pifont}%
\newcommand{\cmark}{\ding{51}}%
\newcommand{\xmark}{\ding{55}}%

\usepackage[pagebackref=true,breaklinks=true,letterpaper=true,colorlinks,bookmarks=false]{hyperref}

\iccvfinalcopy %

\newif\ifsuppmat
\suppmatfalse
\suppmattrue %

\def\iccvPaperID{8564} %

\ificcvfinal\fi

\begin{document}

\title{Warp-Refine Propagation: Semi-Supervised Auto-labeling via Cycle-consistency}

\newcommand{\authorlist}[1][]{%
Aditya Ganeshan\textsuperscript{1}{#1}\quad
Alexis Vallet\textsuperscript{2}\quad
Yasunori Kudo\textsuperscript{2}\quad
Shin-ichi Maeda\textsuperscript{2}\quad
Tommi Kerola\textsuperscript{2}\\
Rare\cb{s} Ambru\cb{s}\textsuperscript{3}\quad
Dennis Park\textsuperscript{3}\quad
Adrien Gaidon\textsuperscript{3}\\
\textsuperscript{1}Brown University\quad
\textsuperscript{2}Preferred Networks, Inc.\quad
\textsuperscript{3}Toyota Research Institute (TRI)
}
\author{
\authorlist[\thanks{Work done while A. Ganeshan was at Preferred Networks, Inc.}]{}
}

\maketitle

\begin{abstract}
Deep learning models for semantic segmentation rely on expensive, large-scale, manually annotated datasets. Labelling is a tedious process that can take hours per image. Automatically annotating video sequences  by propagating sparsely labeled frames through time is a more scalable alternative. In this work, we propose a novel label propagation method, termed Warp-Refine Propagation, that combines semantic cues with geometric cues to efficiently auto-label videos. Our method learns to refine geometrically-warped labels and infuse them with learned semantic priors in a semi-supervised setting by leveraging cycle-consistency across time. We quantitatively show that our method improves label-propagation by a noteworthy margin of 13.1 mIoU on the ApolloScape dataset. Furthermore, by training with the auto-labelled frames, we achieve competitive results on three semantic-segmentation benchmarks, improving the state-of-the-art by a large margin of 1.8 and 3.61 mIoU on NYU-V2 and KITTI, while matching the current best results on Cityscapes.
\end{abstract}

\let\standardclearpage\clearpage

\vspace{-1.5em}
\section{Introduction}
\label{section-intro}
\let\clearpage\relax
\input{sub_2/introduction}

\section{Related Work}
\label{section-rel}
\let\clearpage\relax
\input{sub_2/rel_works}

\section{Warp-Refine Propagation}
\label{section-method}
\let\clearpage\relax

\input{subsections/method}

\section{Experiments}
\label{section-exp}
\let\clearpage\relax
\input{sub_2/experiments}

\section{Conclusion}
\label{section-conclusion}
\let\clearpage\relax
\input{sub_2/conclusions}

\let\clearpage\standardclearpage
\clearpage
{\small
\typeout{}
\bibliographystyle{ieee_fullname}
\bibliography{refs}
}

{\ifsuppmat
\clearpage

\setcounter{section}{1}
\setcounter{table}{0}
\setcounter{figure}{0}
\setcounter{equation}{0}
\renewcommand{\thetable}{\Alph{table}}
\renewcommand{\thefigure}{\Alph{figure}}
\renewcommand{\theequation}{\Alph{equation}}
\appendix

\input{subsections/supp}

}

\end{document}

%% file: sub_2/introduction.tex
Semantic segmentation, i.e. assigning a semantic class to each pixel in an input image, is an integral task in understanding shapes, geometry, and interaction of components from images. The field has enjoyed revolutionary improvements thanks to deep learning \cite{fcn_cvpr, pspnet, unet}. However, obtaining a large-scale dataset with pixel-level annotations is particularly expensive: for example, labeling takes 1.5 hours on average per image in the Cityscapes dataset~\cite{cs_dataset}. Despite the recent introduction of datasets that are significantly larger than their predecessors~\cite{cs_dataset, as_dataset, mscoco, mapillary}, scarcity of labeled data remains a bottleneck when compared to other recognition tasks in computer vision \cite{faster_rcnn, hr_net_pami, roi10d}.
\let\clearpage\relax

\input{figures/figure_miou_prop_apollo}
In the common scenario where data is provided as videos with labels for sparsely subsampled frames, a prominent way to tackle data scarcity is \emph{label propagation} (LP), which automatically annotates additional video frames by propagating labels through time~\cite{lp_2006, lp_2010}. This intuitive idea to leverage motion-cues via temporal consistency in videos has been widely explored, using estimated motion~\cite{lp_2013, sem_warp, lp_eccv}, patch matching~\cite{lp_2010, lp_iccvw}, or predicting video frames~\cite{nvidia_cvpr19}. However, as discussed in Zhu et al.~\cite{nvidia_cvpr19}, estimating dense motion fields across long periods of time remains notoriously difficult. Further, these methods are often sensitive to hyperparameters (e.g. patch size), cannot handle de-occlusion, or require highly accurate optical flow, thus limiting their applicability. %

Another promising approach for obtaining large-scale annotation in semi-supervised settings is \emph{self-training} (ST), in which a teacher model, trained to capture semantic cues, is used to generate additional annotations on unlabeled images \cite{domain_seg_2, domain_seg_10, domain_seg_5, domain_seg_8}. While there have been significant improvements in ST, various challenges still remain in controlling the noise in pseudo-labels, such as heuristic decisions on confidence thresholds~\cite{tao2020hierarchical}, class imbalance in pseudo-labels~\cite{cut_paste}, inaccurate predictions for small segments, and misalignment of category definition between source and target domain. 
\let\clearpage\relax
\input{figures/figure_example}
To mitigate the drawbacks of LP and ST, we propose \emph{Warp-Refine Propagation} (referred to as \emph{warp-refine}), a novel method to automatically generate dense pixel-level labels for raw video frames. 
Our method is built on two key insights: (\romannum{1}) by combining motion cues with semantic cues, we can overcome the respective limitations of LP and ST, and (\romannum{2}) by leveraging \emph{cycle-consistency} across time, we can learn to  combine these two complementary cues in a \emph{semi-supervised} setting without sequentially-annotated videos.

Specifically, our method first constructs an initial estimate by directly combining labels generated via motion cues and semantic cues. This initial estimate, containing erroneous conflict resolution and faulty merges, is then rectified by a separate refinement network.
The refinement network is trained in a \emph{semi-supervised} setting via a novel \emph{cycle-consistency} loss. This loss compares the ground-truth labels with their cyclically propagated version created by propagating the labels forward-and-backward through time in a cyclic loop ($t\rightarrow t + k \rightarrow t$). Our loss is built on the observation that as our auto-labeling method is bi-directional, it can be used to generate different versions of each annotated frame. Once this network is trained, it is used to correct errors caused by propagation of variable length. In Fig.~\ref{fig:window-in-window} we show a qualitative comparison of our method against prior-arts, demonstrating drastic improvements in label quality.

With quantitative analysis on a large scale autonomous driving datasets (ApolloScape~\cite{as_dataset}), we concretely establish the superior accuracy of our method against previous \emph{state-of-the-art} auto-labeling methods. Such an analysis of different methods has been starkly missing from prior works~\cite{nvidia_cvpr19, tao2020hierarchical, lp_eccv}. As shown in Fig.~\ref{fig_miou_time}, we observe that \emph{warp-refine} accurately propagates labels for significantly longer time intervals, with a notable average improvement of $13.1$ \textit{mIoU} over the previous best method on ApolloScape. Further, it accurately labels rare classes such as `Bicycle' and thin structures such as `Poles' (cf. Section~\ref{subsec:evaluating_propagated_labels}). As a result, by training single-frame semantic segmentation models with the additional data labeled by our method, we achieve \textit{state-of-the-art} performance on KITTI~\cite{kitti_dataset}, NYU-V2~\cite{nyu_dataset} and Cityscapes~\cite{cs_dataset} benchmarks (cf. Section~\ref{subsec-sota}).

In summary, our main contributions are: 1) A novel algorithm, termed \emph{Warp-Refine Propagation}, that produces significantly more accurate pseudo-labels, especially for frames distant in time; 2) A novel loss function, based on the \emph{cycle-consistency} of learned transformations, to train our method in a \emph{semi-supervised} setting; and 3) A quantitative analysis on the \emph{quality} and \emph{utility} of different auto-labeling methods on multiple diverse datasets. To the best of our knowledge, our work is the first to utilize both semantic and geometric understanding for the task of video auto-labeling.

%% file: figures/figure_miou_prop_apollo.tex
\begin{figure}[t]
	\centering
	\includegraphics[width=1.00\linewidth]{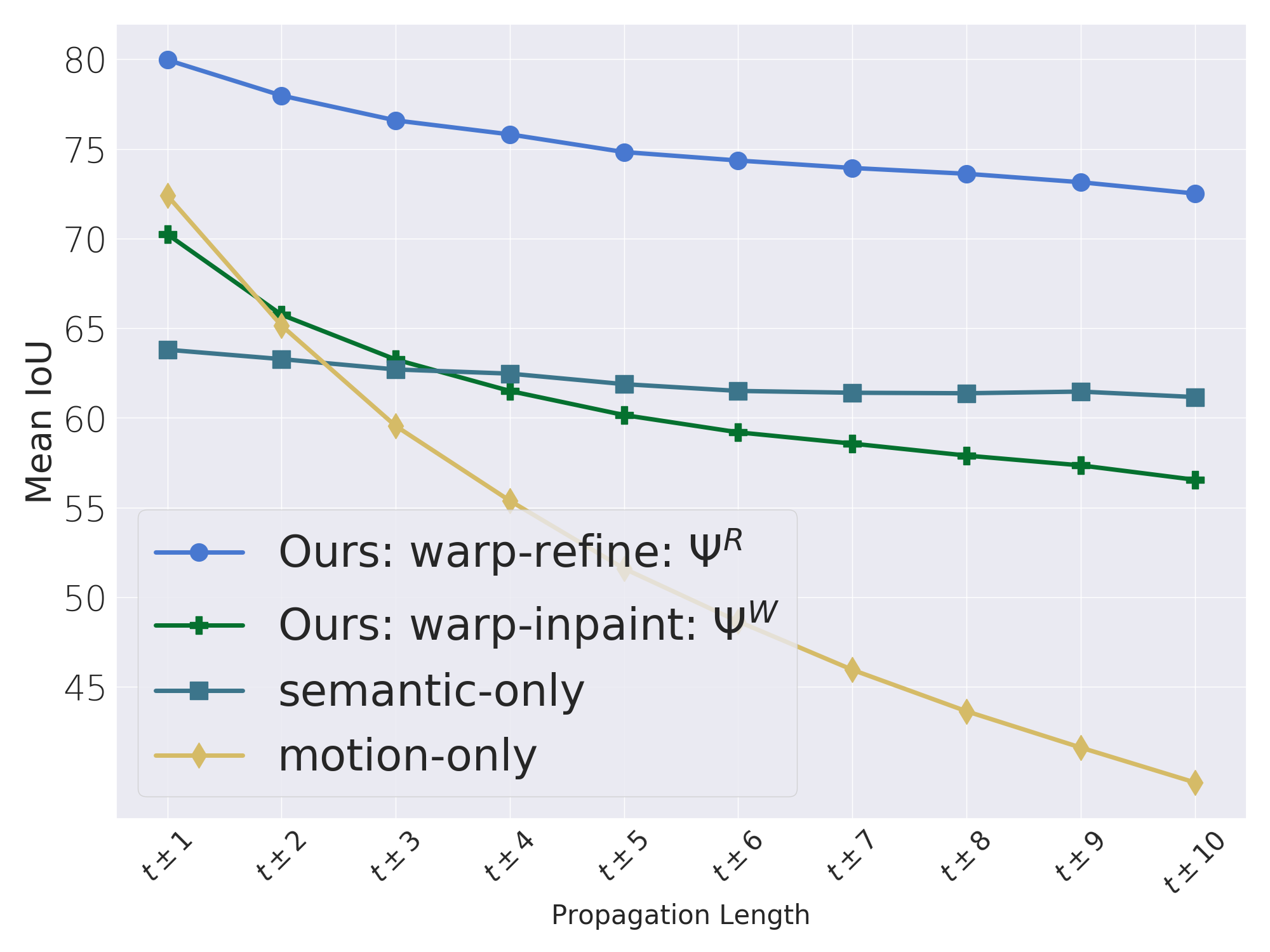}
	\caption{\small Quantitative comparison of different auto-labelling methods for multiple propagation lengths using ApolloScape~\cite{as_dataset} frame-wise ground-truth. We compare our propagation methods, \emph{warp-refine} and \emph{warp-inpaint}, to prior-arts \textit{semantic-only} and \textit{motion-only} propagation, and show that  \emph{warp-refine} is vastly superior to other auto-labeling approaches, especially for large time-steps.}

	\label{fig_miou_time}
    \vspace{-2ex}
\end{figure}

%% file: figures/figure_example.tex
\begin{figure*}[t]
	\centering
	\includegraphics[width=1.0\linewidth]{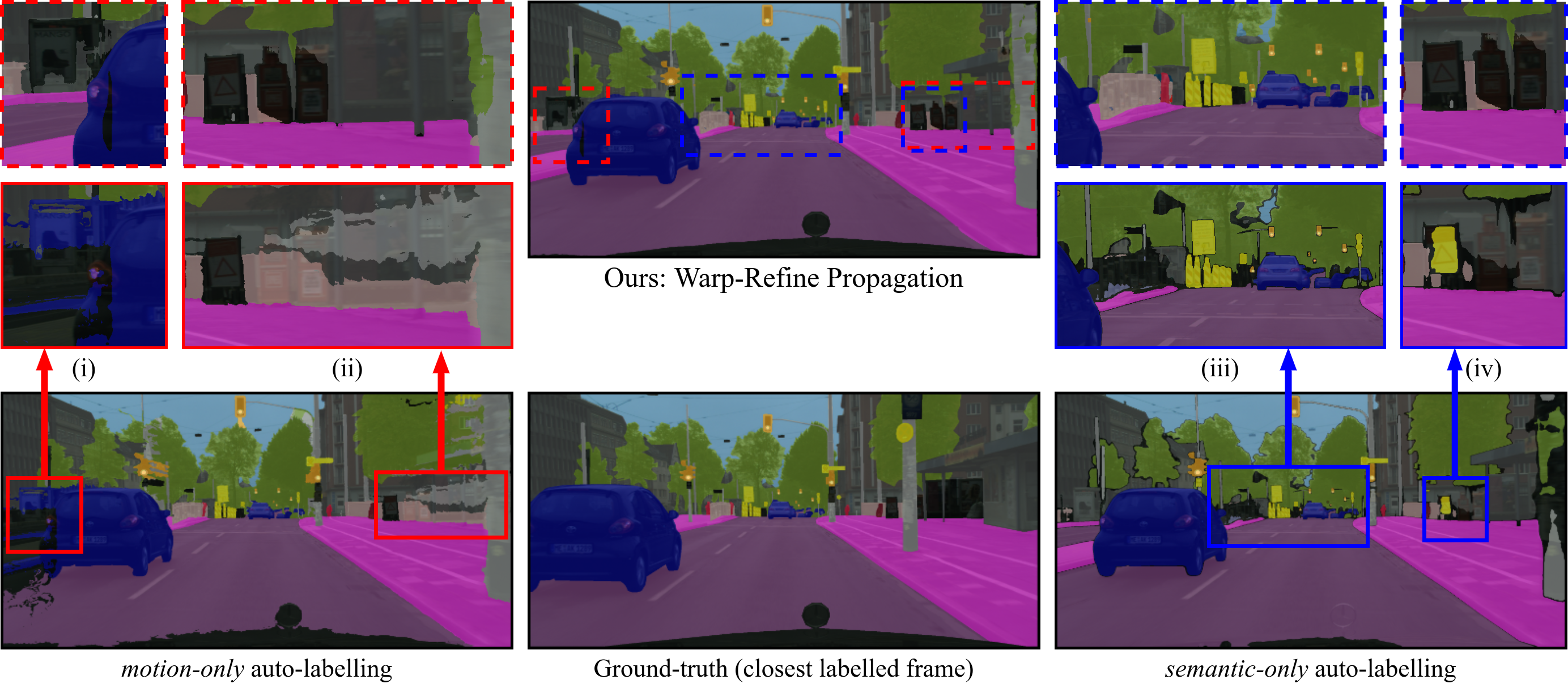}
	\caption{{\small Accuracy of propagated labels. We visually compare the proposed \emph{warp-refine} propagation (\textbf{top-center}) with the \emph{motion-only}  model (\textbf{bottom-left}), the \emph{semantic-only} model (\textbf{bottom-right}), and the ground-truth annotation (\textbf{bottom-center}). The \emph{motion-only} model (\romannumeral 1) fails to correctly classify the new regions introduced in the target frame , and (\romannumeral 2) often suffers from \emph{drifting} . In contrast, the semantic-only model (\romannumeral 3) tends to fail for far-away segments, and (\romannumeral 4) cannot handle misaligned class definitions between the teacher and the student model (e.g. \emph{ignore} label) . Our method effectively combines the strengths of both of these approaches to overcome their respective limitations. For details of \emph{motion-only} and \emph{semantic-only} models, see Section~\ref{subsec:ablative}.}
	} 
	
	\label{fig:window-in-window}
    \vspace{-1.0ex}
\end{figure*}

%% file: sub_2/rel_works.tex
\let\clearpage\relax
\input{figures/figure_proposed_method}

\noindent
\textbf{Self-training (ST).} The approach of applying a mature teacher networks to unlabeled images and using the predicted labels to supervise student networks has received increasing attention. Xie et al.~\cite{noisy_self} introduces a framework for ST for controlling noise in pseudo-labels to exceed the teacher network. Chen et al.~\cite{google_student} extend it for semantic segmentation. Recently, ST has proved to be effective on unsupervised domain adaptation (UDA)~\cite{domain_seg_5, domain_seg_7, domain_seg_8, domain_seg_9}, where the goal is to learn a domain invariant representation to address the \emph{sim-to-real} gap. The advances in ST are mainly driven by improvements in model architectures and loss definitions used in training the teacher and student networks~\cite{fcn_cvpr, rcnn_1, deeplab_v1, deeplab_v2, pspnet, unet}, feature alignment between source and target domain~\cite{domain_seg_9, domain_seg_nips_2, domain_seg_2}, and decision process of which predictions are used as pseudo-labels~\cite{domain_seg_1, domain_seg_5, domain_seg_7, domain_seg_8}. 
Our approach can be seen as further improving ST by 1) additionally using motion cues to propagate ground-truth labels across long intervals in time and 2) correcting the combined errors from the two sources using learnable components.

\noindent
\textbf{Label propagation (LP).} The goal of LP is to transfer ground-truth pixel labels from adjacent frames using dense pixel-level matching~\cite{lp_eccv, lp_iccvw, lp_2013, nvidia_cvpr19}. Given the advance in CNN-based optical flow algorithms, various methods using strong geometric cues to propagate labels have been proposed, \eg using video prediction~\cite{nvidia_cvpr19} and flow-based voting~\cite{Marcu_2020_ACCV}. 
The key criteria for success in LP is the \emph{distance in time} through which one can accurately propagate ground-truth labels. It is a vital aspect, as the pseudo-labels need to contain novel learning signals with respect to the annotated source frames. A common failure of flow-based methods is error propagation; mistakes made on earlier steps persist and get amplified in later steps (i.e. \emph{drifting}). 
Here, our method can be seen as improving LP by using a high-capacity semantic segmentation model to \emph{re-initialize} the labels to prevent error accumulation.

\noindent
\textbf{Semantic and geometric cues.} Methods combining semantic and geometric cues have been proposed in the past for other tasks such as future-frame prediction~\cite{future_seg} and video-segmentation\cite{sem_warp, feelvos2019, prop_rt}. Our chief distinction is the propagation of ground-truth labels, not feature representation or predicted labels, for usage in a self-training framework.
In Sec.~\ref{subsec:ablative}, we provide quantitative analysis and report that our method is successful in propagating ground-truth labels accurately across long intervals of time.

\vspace{1em} %
\noindent
\textbf{Cycle-consistency.} The concept of cycle-consistency has been previously utilized for learning object embeddings~\cite{CVPR2019_CycleTime}, one-shot semantic  segmentation~\cite{Wang_2020_CVPR} and video interpolation~\cite{cycle_vid_interp}. Our work is inspired by work using cycle-consistency for learning a robust tracker~\cite{CVPR2019_CycleTime}. However, we differ in that we address the noisy nature of our tracking/geometric modelling method itself.

%% file: figures/figure_proposed_method.tex
\begin{figure*}[t]
	\centering
	\includegraphics[width=1.00\linewidth]{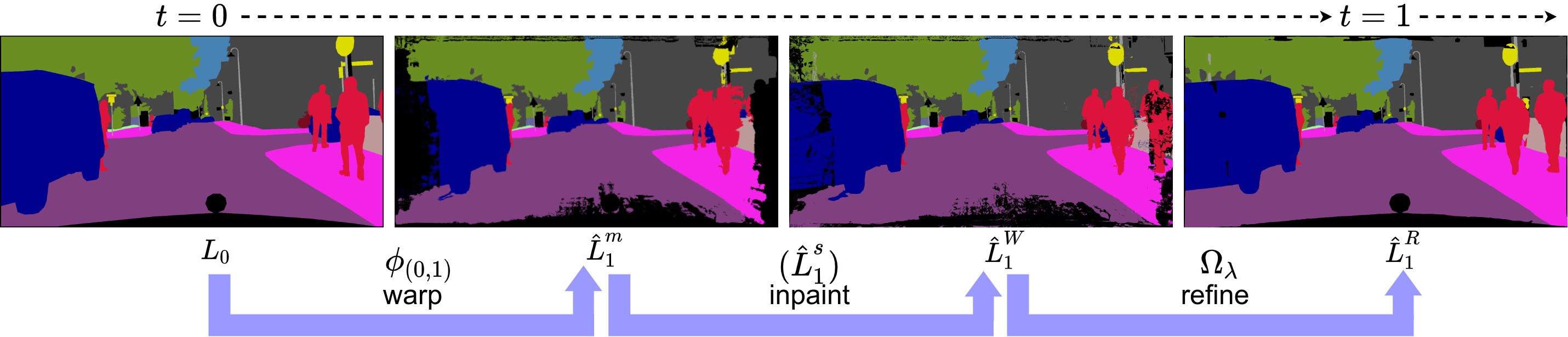}
	\caption{\small Our proposed \textit{warp-refine} propagation consists of three steps: \textbf{(i)} \textbf{warp} uses dense optical flow estimation to remap ground-truth labels onto target images; \textbf{(ii)} \textbf{inpaint} blends the results of (i) with predictions of a strong semantic segmentation model; \textbf{(iii)} \textbf{refine} applies the learned refinement network on the results of (ii).}
	\label{fig:proposed-method}
\end{figure*}

%% file: subsections/method.tex
We first present the notation used throughout our paper. This is followed by the description of two recursive algorithms for propagating dense pixel labels, followed by the proposed method to train the denoising models by leveraging cyclic consistency.

\subsection{Notation}

Given a labeled video frame $(I_t, L_t)$ and its adjacent raw frame $I_{t+k^{'}}$, we aim to create approximated labels $\hat{L}_{t+k{'}}$ for $1\leq k^{{'}} \leq K$. To this end, we introduce two greedy propagation algorithms. They are greedy in that the optimal solution for $\hat{L}_{t+k^{'}}$, namely $\hat{L}^*_{t+k{'}}$, is obtained by applying a recursive propagation step to the (approximately optimal) solution for the previous frame, $\hat{L}^*_{t+k^{'}-1}$:\footnote{To avoid clutter in subscripts, we define $k \coloneqq t + k^{'}$.}
\begin{align}
    \hat{L}^*_{k} &= \Psi(\hat{L}^*_{k-1}, I_{k-1})~, \quad t+1\leq k\leq t+K~, \\
    \hat{L}^*_t &= L_t~.
\end{align}
We introduce two algorithms, \textbf{warp-inpaint} and \textbf{warp-refine}, that grow in complexity of $\Psi$. Notably, the approach of Zhu et al.~\cite{nvidia_cvpr19} can be included in this framework, when only video-prediction algorithm~\cite{sdc_net} based motion vectors are used for defining $\Psi$. 

\subsection{Warp-Inpaint}
\label{subsection:warp-inpaint}
As commonly observed in online visual tracking~\cite{ilg2017flownet}, when $\Psi$ relies purely on motion cues, the propagated labels are susceptible to propagation error (i.e. \emph{drifting})~\cite{nvidia_cvpr19}. In addition, the pixels of new scene elements cannot be explained by labels in previous frames (e.g. cars entering the field of view). One way to address this is to \emph{re-initialize} the semantic labels using a strong semantic segmentation model.

We therefore allow each pixel in $\hat{L}^*_{k}$ to be derived either from \emph{motion} cues encoded in the $(I_{k-1}, I_k)$ pair \textbf{or} from \emph{semantic} cues computed solely from $I_k$. Formally, we compute a version of $\hat{L}_k$, namely $\hat{L}^m_k$, by remapping $\hat{L}^*_{k-1}$ using a transformation $\phi_{k-1, k}$ learned to warp $I_{k-1}$ onto $I_k$. We then blend $\hat{L}^m_k$ with another version of $\hat{L}_k$, namely $\hat{L}^s_k$, which are semantic labels obtained by applying a pretrained semantic segmentation model $g_\psi$ on $I_k$:
\begin{align}
\hat{L}^m_k &= \phi_{k-1, k}(\hat{L}^*_{k-1})~, \label{eq:warp-start} \\
\hat{L}^{s}_{k} &= g_\psi(I_k)~, \\
\hat{L}^*_k &= M \odot \hat{L}^m_k + (1 - M) \odot \hat{L}^s_k~,\label{eq:warp-end}
\end{align}
where $\odot$ denotes pixel-wise multiplication. The $(x, y)$ value of the binary mixing coefficients $M$ represents whether we trust the the estimated motion vector at the position $(x, y)$, compared with the semantic label computed by $g_\psi$. We determine $M$ by measuring the Euclidean distance of pixel values between $I_k$ and $\phi_{k-1, k}(I_{k-1})$:
\begin{align}
    M(x, y) &= \mathbb{I}(\|I_{k}(x, y) - \phi_{k-1, k}(I_t)(x, y)\|_2 < \tau)~,
\end{align}
\noindent where $\mathbb{I}(\cdot)$ is the indicator function. The motion vectors are obtained by applying a pretrained motion estimation model, $f_\theta$, to neighboring image pairs: $\phi_{k-1, k} = f_\theta(I_{k-1}, I_k)$.
We let $\Psi^W$ denote the entire propagation process \eqref{eq:warp-start} - \eqref{eq:warp-end}, and $\hat{L}_k^W$ denote the resulting pseudo-labels at the $k$-th frame:
\begin{align}
    \hat{L}_k^W &= \Psi^W(\hat{L}_{k-1}^W, I_{k-1})~.
\end{align}

\subsection{Warp-Refine}
\label{subsection:warp-refine}
With $\hat{L}^W_k$, we obtain an initial fusion of geometric and semantic cues. This estimate is however still subject to artifacts from imperfect motion estimation and semantic segmentation models ($f_\theta$ and $g_\psi$). We therefore extend the recursive step to refine $\hat{L}^W_k$ by applying a de-noising network $\Omega_\lambda$ that aims to remove these artifacts:
\begin{align}
\hat{L}^R_k &= \Omega_\lambda(\hat{L}^{W}_k)~.
\end{align}
Note that the goal of $\Omega_\lambda$ is to mitigate particular types of errors caused by the propagation steps of $\Psi^W$,  specifically by $f_\theta$, $g_\psi$, and the choice of $\tau$, and in effect, properly merge the semantic and geometric cues.
The extended propagation process and generated pseudo-labels are denoted by $\Psi^R$ and $\hat{L}^R_k$, respectively: $\hat{L}_k^R = \Psi^R(\hat{L}_{k-1}^R, I_{k-1})$.

Fig~\ref{fig:proposed-method} summarizes the three steps for our proposed propagation method.
In Sec.~\ref{section-exp}, we quantitatively and qualitatively show that $\hat{L}^W_k$ as well as $\hat{L}^R_k$ are more accurate with respect to their ground truth labels for various $k$, and further that the generated labels are useful for improving single-frame semantic segmentation models.

\let\clearpage\relax
\input{figures/figure_training}

\subsection{Learning}
There are three sets of learnable parameters in \textit{warp-refine}; in the motion estimation model $f_\theta$, in the semantic segmentation model $g_\psi$, and in the de-noising model $\Omega_\lambda$. We use a constant pretrained motion estimation model~\cite{sdc_net} and semantic segmentation model~\cite{tao2020hierarchical, ocr_eccv_20} for $f_\theta$ and $g_\psi$, respectively, and use a fixed value for $\tau$. Here we describe how to train $\Omega_\lambda$.

\textbf{Cycle-consistency.} Training a de-noising model in a fully-supervised setting typically requires a large dataset of \emph{noisy-clean} pairs~\cite{isola2017image}, which in our case is $(\hat{L}^R_k, L_k)$. To address the lack of $L_k$ in our semi-supervised setting, we leverage the cycle-consistency inherent in our propagation mechanism. The cyclic propagation consists of two stages:
\begin{enumerate}
  \item \emph{forward}: we execute propagation steps of $\Psi^W$ (i.e. Eq.~\ref{eq:warp-start} - Eq.~\ref{eq:warp-end}) $l$ times to obtain $\hat{L}^{W}_l$. 
  \item \emph{backward}: we execute $l$ steps of \emph{inverted} propagation of $\Psi^W$ to obtain a \emph{cyclically propagated} variant of $L_t$, namely $\hat{L}^{\circ}_t$.
\end{enumerate}
The inverted propagation step is similar to $\Psi^W$, but executed in a reverse order:
\begin{align}
\hat{L}^m_{k-1} &= \phi_{k, k-1}(\hat{L}^{*}_k)~, \\
\hat{L}^{s}_{k-1} &= g_\psi(I_{k-1})~, \\
\hat{L}^{*}_{k-1} &= M \odot \hat{L}^m_{k-1} + (1 - M) \odot \hat{L}^s_{k-1}~, 
\end{align}
\noindent where $\hat{L}^*_k$ is set to the result of the forward stage, $\hat{L}^{W}_l$, and motion vectors are computed in backward:
\begin{align}
    \phi_{k, k-1} &= f_\theta(I_k, I_{k-1})~. 
\end{align}
The cycle-consistency loss is computed by comparing the annotated labels $L_t$ and its cyclically warped counterpart $\hat{L}^\circ_t$ with de-noising applied: $\mathcal{L}(L_t, \Omega_\lambda(L^\circ_t))$. This  is optimized via standard gradient-based methods during training.

Notably, the backward steps, and therefore the entire forward-backward process, chains the same set of transformations used in $\Psi^W$ (i.e. $f_\theta$, $g_\psi$ and the blending strategy). Therefore, the de-noising network trained with this cycle-consistency loss is expected to correct the errors in pseudo-labels generated by variable length of the $\Psi^W$, which is the goal of $\Omega_\lambda$. In practice, we use varying $l$ in training to maximize this generalizability. 
Fig.~\ref{fig:training} presents a visual summary of our approach for learning label refinement via cycle-consistency of labels.

%% file: figures/figure_training.tex
\begin{figure}[t]
	\centering
	\includegraphics[width=0.95\linewidth]{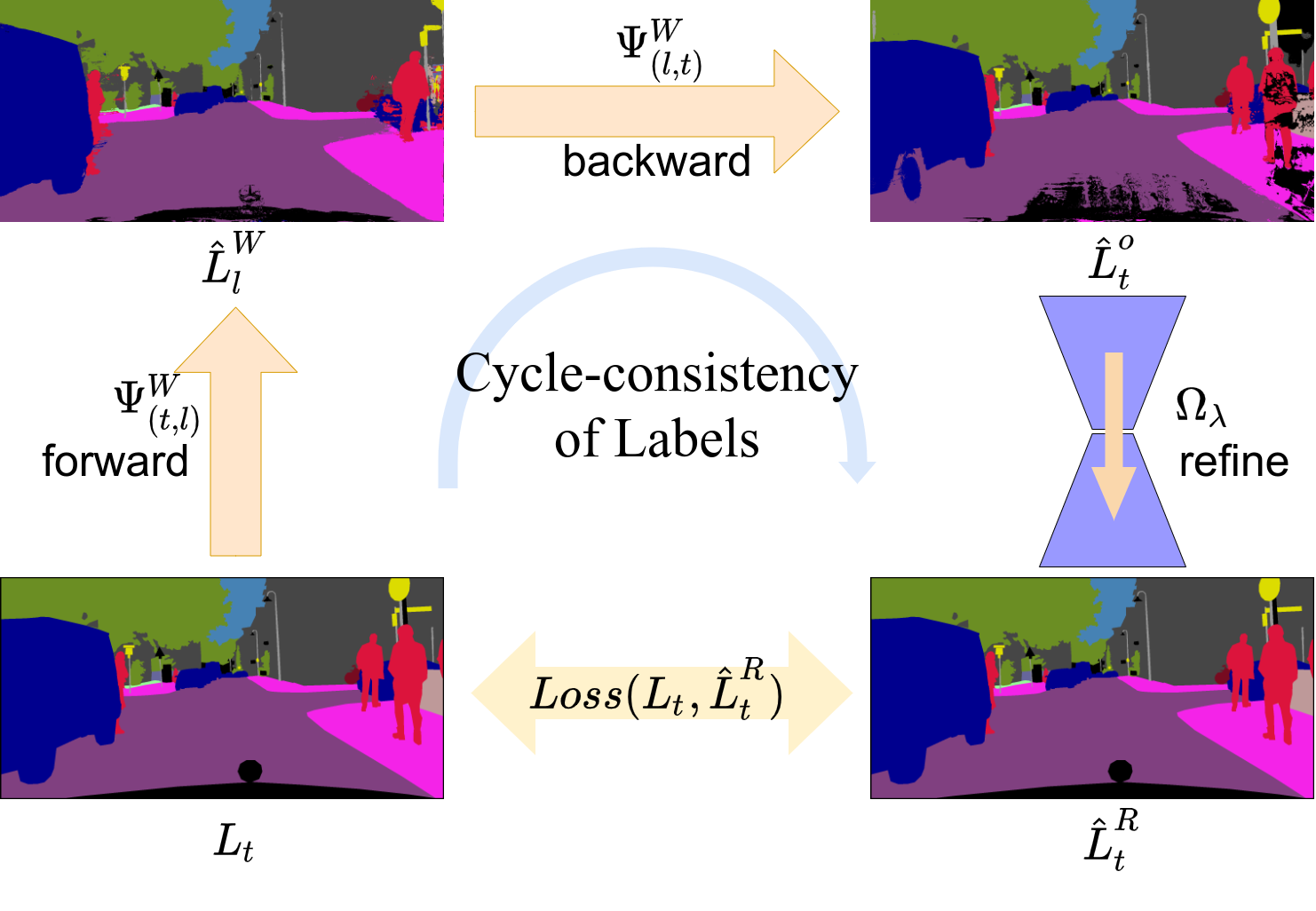}
	\caption{\small For training the refinement network, we generate cyclically propagated labels by applying forward-backward propagation on annotated frame using \emph{warp-inpaint} transformation, and compare them with the ground-truth labels.}
	\label{fig:training}
\end{figure}

%% file: sub_2/experiments.tex
\subsection{Datasets} 
\label{subsec:datasets}
\vspace{-0.5em}
We present quantitative and qualitative experiments on four semantic segmentation benchmarks: \\
\textbf{NYU.} The NYU-Depth V2 dataset~\cite{nyu_dataset} consists of 1449 densely labeled images split into 795 training images and 654 testing images, which are randomly sampled from video sequences.  Due to the high fps (20-30 fps) and slow camera movement, we sub-sample the video at 2fps and create sequences of variable lengths of up to 21 frames around the labeled frame, yielding $9786$ frames for label propagation.\\
\textbf{KITTI.} The KITTI Vision Benchmark Suite~\cite{kitti_dataset} consists of 200 training and 200 test images with pixel-level annotation. For each of the training images, we use sequential frames ($\pm 10$) from the scene-flow subset for label propagation. \\
\textbf{Cityscapes.} The Cityscapes dataset~\cite{cs_dataset} is split into a train, validation, and test set with 2975, 500, and 1525 images, respectively. For each training image, we use the  $\pm 10$ unlabeled neighboring frames provided as part of the dataset. \\
\textbf{ApolloScape.} The ApolloScape dataset~\cite{as_dataset} contains pixel-level annotations for sequentially recorded images, divided as 40960 training, and 8327 validation images, allowing evaluation of the accuracy of the propagated labels. We create continuous partitions of 21 frames, and use the central frame as a training data-point, and the adjacent frames ($\pm 10 $) for label propagation. This yields a train-subset of size containing 2005 and a label-propagation subset of 38095 images.

\vspace{-0.5em}
\subsection{Implementation Details}
\label{subsec:implementation}
\vspace{-0.5em}

Our approach consists of three components: a motion-estimation network $f_\theta$, the semantic segmentation model $g_\psi$, and a de-noising network $\Omega_\lambda$. Following Zhu et al.~\cite{nvidia_cvpr19}, for the motion-estimation network $f_\theta$, we use video-prediction model based on SDC-net~\cite{sdc_net}. For our task, video prediction performed better than warping with optical-flow~\cite{raft_of} (cf. supplementary). For segmentation model $g_\psi$, we adopt the architecture (MSA-HrNet-OCR) and training protocols outlined in Tao et al.~\cite{tao2020hierarchical}. 
Finally for $\Omega_\lambda$, we use a pix2pix-style network ~\cite{pix2pix, spades}. First, an encoder takes the warp-inpainted labels as the input. The encoding formed is then concatenated with OCR features~\cite{ocr_eccv_20} from $g_\psi$, and passed through a decoder.
For robust refinement, we perform cycle-consistency training with the range of propagation sampled till $\pm6$. Note that the labels ($L_t\text{/}L^W_t\text{/}L^\circ_t$) are utilized as one-hot vectors.

\noindent
\textbf{Single-frame semantic segmentation.}~For training single frame semantic segmentation networks with the auto-labeled data, we use the same architecture and training protocol as followed for $g_\psi$, with one important modification: independent of the amount of data generated by auto-labeling, we use a fixed epoch size (thrice the dataset size). This ensures that we do not under-train the baseline model (cf. supplementary). When training with the propagated data, we sample  $70\%$ of the epoch from the propagated data, and $30\%$ from the manually annotated data. Unless specified otherwise, we sample additional data at time-frames $\{t \pm n| n \in [2,4,6,8]\}$ (similar to ~\cite{lp_eccv}). We refer to the model trained with no additional data as the \emph{baseline} model.

\noindent 
\textbf{Auto-labeling baselines.}~We compare our methods \emph{warp-inpaint} ($\Psi^{W}$, cf. section~\ref{subsection:warp-inpaint}) and \emph{warp-refine} ($\Psi^{R}$, cf. section~\ref{subsection:warp-refine}) against existing auto-labeling techniques. We use the method proposed by Zhu et al.~\cite{nvidia_cvpr19}, which uses only $f_\theta$ to generate the labels, and refer to this method as \textit{motion-only} labeling. When using these labels, we use joint image-label propagation as recommended by~\cite{nvidia_cvpr19}. Similarly, we also use the method proposed by Tao et al.~\cite{tao2020hierarchical}, which generates labels using only $g_\psi$, and refer to this method as \textit{semantic-only} labeling. For \textit{semantic-only} labeling, we use the best performing architecture (MSA-HRNet-OCR, trained on the manually annotated images), and use only the pixels with $>0.9$ confidence, as recommended in ~\cite{tao2020hierarchical}.

\subsection{Quality of Propagated Labels}
\label{subsec:evaluating_propagated_labels}

We first provide extensive analysis of the our auto-labeling methods: \textit{warp-inpaint} and \textit{warp-refine}, and existing techniques \textit{motion-only} and \textit{semantic-only}. We evaluate the labels generated by these methods against the ground-truth labels provided in the ApolloScape dataset~\cite{as_dataset}. We focus on two crucial aspects: (1) long-range propagation, and (2) labeling of hard classes.

\let\clearpage\relax

\input{figures/figure_miou_per_class}

First, we compare the different auto-labeling methods across various propagation lengths. Fig.~\ref{fig_miou_time} reports the mean Intersection over Union (\emph{mIoU}) between propagated labels and ground-truth labels, given various propagation methods and propagation lengths.
We note that the \emph{motion-only} and the \emph{semantic-only} models show a clear trade-off with respect to the propagation length: \emph{motion-only} model produces more accurate labels over shorter ranges, while \emph{semantic-only} model over longer ranges.
Further, when propagating without refinement (i.e. \textit{warp-inpaint}), the accuracy degrades for longer propagation lengths, even dropping below the \emph{semantic-only} labeling. Finally, propagating with refinement (i.e. \emph{warp-refine}) produces significantly cleaner labels than the others. Due to the refinement module, our method retains its accuracy even at large time-steps, attaining a large margin of $11.35$ \textit{mIoU} over the closest competing method at $t \pm 10$.

Next, we quantify the \textit{IoU} of difficult classes and the overall \textit{mIoU} across all propagation lengths in Fig.~\ref{fig:iou_per_class}.  Notably, both prior methods for auto-labeling fail on thin structures such as `Poles' - \textit{motion-only} causes over-labeling due to drifting, while \textit{semantic-only} causes under-labeling due to low-confidence predictions (such regions are frequently masked out). Further, we note that \textit{semantic-only} labeling severely fails at estimating the `Ignore' class, in contrast to \textit{motion-only} labeling which yields accurate estimates. As the `Ignore' class consists of widely varying objects, it is difficult to model it as a semantic class (typically `Ignore' class includes regions labeled as `Others' and `Unlabeled'; i.e. labels which do not have any semantic definition). Therefore, following the recommended protocol~\cite{tao2020hierarchical}, for \textit{semantic-only} labeling we estimate `Ignore' regions via probability thresholding~\cite{tao2020hierarchical} (cf. Sec.~\ref{subsec:implementation}). Despite the careful selection of this threshold parameter, \textit{semantic-only} labeling fails to accurately label the `Ignore' class. Our methods (\textit{warp-inpaint}, and \textit{warp-refine}) effectively combine motion-cues with semantic-cues, thus overcoming this drawback, and properly estimate the `Ignore' class. Overall, our method again yields an impressive margin of $13.12$ \textit{mIoU} over the closest prior-art (averaged across all propagation lengths).

\let\clearpage\relax

\input{figures/figure_qualitative_as}

Finally, we also present qualitative results in Fig.~\ref{fig_as_qualitative}, noting that our propagated labels do not suffer from error propagation compared to \textit{motion-only}, while achieving higher accuracy on the rare classes (e.g. bikes) compared to \textit{semantic-only}. Fig.~\ref{fig_cs_qualitative} also presents a failure case for our method, caused due to the concurrent failure of both semantic and motion cues for the `Rider'.

\let\clearpage\relax
\input{tables/nyu_ablation}

\subsection{Utility of Propagated Labels}
\label{subsec:ablative}

We now demonstrate that the significant improvements in auto-labeling directly translate to superior performance for single-frame semantic segmentation models trained with our generated data. We perform our experiments on NYU-V2, ApolloScape and Cityscapes. Following Zhu et al.~\cite{nvidia_cvpr19}, for each experiment, we perform three runs and report the mean ($\mu(\textit{mIoU})$) and sample standard deviation ($\sigma(\textit{mIoU})$). Our analysis is summarised in Table~\ref{table:nyu_ablation}.

\noindent
\textbf{NYU-V2 \& ApolloScape.} Training with \emph{warp-refine} labels consistently yields better results. On NYU-V2, our labels yield an average improvement of $1.54$ \textit{mIoU}, compared to only $0.35$ \textit{mIoU} for the \textit{semantic-only} labeling method. Similarly, on ApolloScape, \textit{warp-refine} labels increase performance by $1.11$ \textit{mIoU}, whereas the closest auto-labeling baseline (\textit{semantic-only}) yields a benefit of only $0.47$. 
Finally, we note that training with \textit{motion-only} labels consistently leads to a drop in performance.

Since ApolloScape contains the ground-truth annotation of the all the provided images, we also provide evaluation using the ground-truth labels instead of the labels generated via auto-labeling. This acts as an \textit{oracle} propagation model, and we treat it as an empirical upper-bound on the benefits from label propagation. Using the ground-truth instead of propagated labels (at $(t\pm \{2,4,6,8\})$) yeilds a benefit of $2.73$ \textit{mIoU}. Notably, training with \textit{warp-refine} labels attains about $40\%$ of this empirical upper-bound. 

\let\clearpage\relax

\input{figures/figure_qualitative_cs}

\noindent 
\textbf{Cityscapes-val.} While \textit{warp-refine} labels yields  performance superior to prior methods, we notice that the gap between \textit{semantic-only} and \textit{warp-refine} is smaller on Cityscapes (while still being statistically significant). It is likely due to performance saturation: across the three datasets, we observe that as the performance of the baseline model increase, (1) the utility of \textit{semantic-only} labels increases (as they are more accurate), and (2) the utility of ~\textit{warp-refine} labels decreases (as labels in adjacent images become less useful). 
Despite this drawback, \textit{warp-refine} labels are still significantly more effective than the prior arts. Labeling additional data via a teacher model has been recently used in many self-training approaches~\cite{noisy_self, google_student}. Our work shows that using geometric cues can further improve the labels, leading to increased performance.

\let\clearpage\relax

\input{figures/figure_ablation_1}

\noindent
\textbf{Propagating over longer time-horizons.} The efficacy of the propagated labels depends on the \textit{how far} the ground-truth labels can be accurately propagated. This is critical as there is little information gain in the immediate neighbors of an annotated frame. 
In Figure~\ref{fig_nyu_seq_size}, we observe that training semantic segmentation models with \textit{warp-refine} labels sampled from long time ranges yields a large benefit. This demonstrates that the ability of \emph{warp-refine} to accurately propagate labels onto remote frames is critical for effectively improving semantic-segmentation models. 
Note that training with \textit{motion-only} labels only degrade the performance further as the propagation length increases. In contrast, our method shows no degradation, even while sampling labels from $[-10, 10]$.

\let\clearpage\relax
\input{tables/table_sota_v2}

\subsection{Semantic Segmentation Benchmarks}
\label{subsec-sota}
Finally, we tabulate the \textit{state-of-the-art} performance achieved by our method on three semantic segmentation benchmarks: NYU-V2, KITTI, and Cityscapes-test. As the evaluation rules for KITTI and Cityscapes explicitly state that the test-split should not be used for ablative study, we only evaluate our final model, i.e. MSA-HrNet-OCR~\cite{tao2020hierarchical} trained with \textit{warp-refine} labels. 
\let\clearpage\relax
\input{tables/nyu_sota}

\noindent
\textbf{NYU-V2.} We compare our model with the best reported scores on the dataset in Table~\ref{tab_nyu_sota}. Our method yields an improvement of $1.8$ \textit{mIoU} over the prior state-of-the-art, while also attaining favorable statistics for the other metrics, notably, an increase of $3.5\%$ in class-wise mean pixel-accuracy (\textit{mean-acc}). Due to the semantic complexity of NYU-V2, additional data is decisively beneficial. Specifically, long-tail classes such as `Bag', `White-board' and `Shower-curtain' yield an average benefit of $7.29$ \textit{IoU} over the baseline.

\noindent
\textbf{KITTI.} We report our performance on the KITTI dataset in Table~\ref{tab_kitti_sota}. We show a significant increase over the previous state-of-the-art. Specifically, we improve by a large margin of $3.61$ \textit{mIoU}. To evaluate the benefits from \textit{warp-refine} labels, we trained our model without \textit{warp-refine} labels, and qualitatively compare the predictions on test-set images (provided in the supplementary). Training with the proposed \textit{warp-refine} labels improves performance on classes such as `Truck', which have only a few labeled examples in the $200$ training images. We attribute the superior performance of our method on this dataset to the fact that adding labeled data is vastly beneficial due to the small size of this dataset. Note that as in Zhu et al.~\cite{nvidia_cvpr19}, we use a pretrained Cityscapes model for initialization, and estimate hyperparameters using 4-fold cross-validation on the train set.

\let\clearpage\relax

\input{tables/kitti_sota}

\noindent
\textbf{Cityscapes-test.} Finally, we explore the benefit of our method on the Cityscapes dataset. As reported in Table~\ref{tab_cs_sota_v2}, our method yields an \textit{mIoU} of $85.3$, improving upon the state-of-the-art by a small margin of $0.1$ \textit{mIoU}. We attribute the small increase to the highly saturated performance of the baseline model. Specifically, as label-propagation only labels the frames neighboring the annotated images, the utility of these pseudo-labels diminishes as the performance of the baseline model gets saturated. We present additional experiments and results with (i) different backbone networks; and (ii) different training regimes in the supplementary.

%% file: figures/figure_miou_per_class.tex
\begin{figure}[t]
	\centering
	\includegraphics[width=0.95\linewidth]{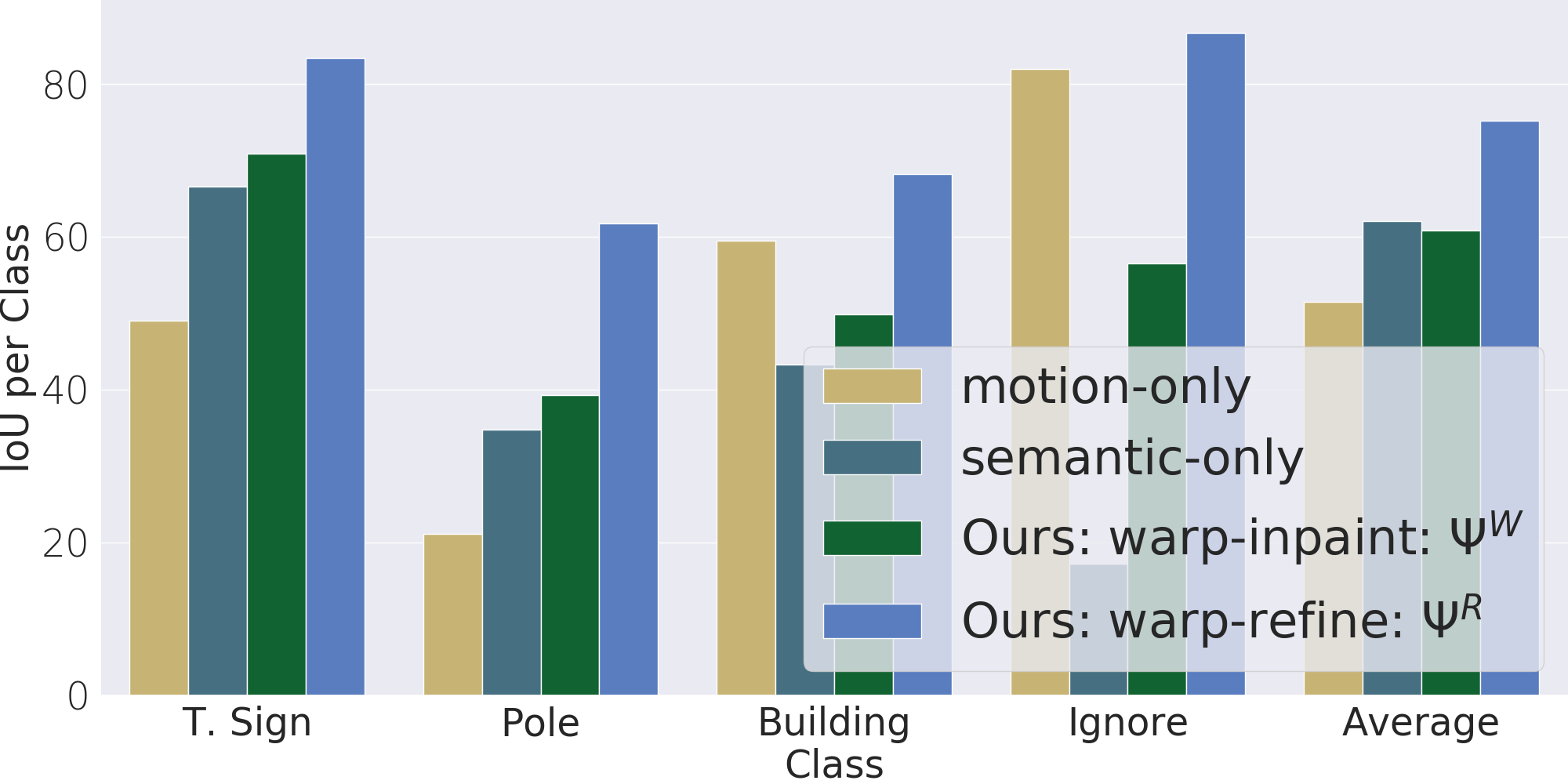}
    \vspace{-1.0em}
	\caption{\small Comparison of propagated labels for difficult classes in ApolloScape~\cite{as_dataset} averaged across all time-steps. Our method \textit{warp-refine} ($\Psi^R$) is able to perform well for classes with thin-structures such as `Pole'. Further, our method leverages motion-cues when semantic-cues fail, as seen for `Ignore' and `Building' class.}
	\label{fig:iou_per_class}
\end{figure}

%% file: figures/figure_qualitative_as.tex
\begin{figure}[t]
	\centering
	\includegraphics[width=0.95\linewidth]{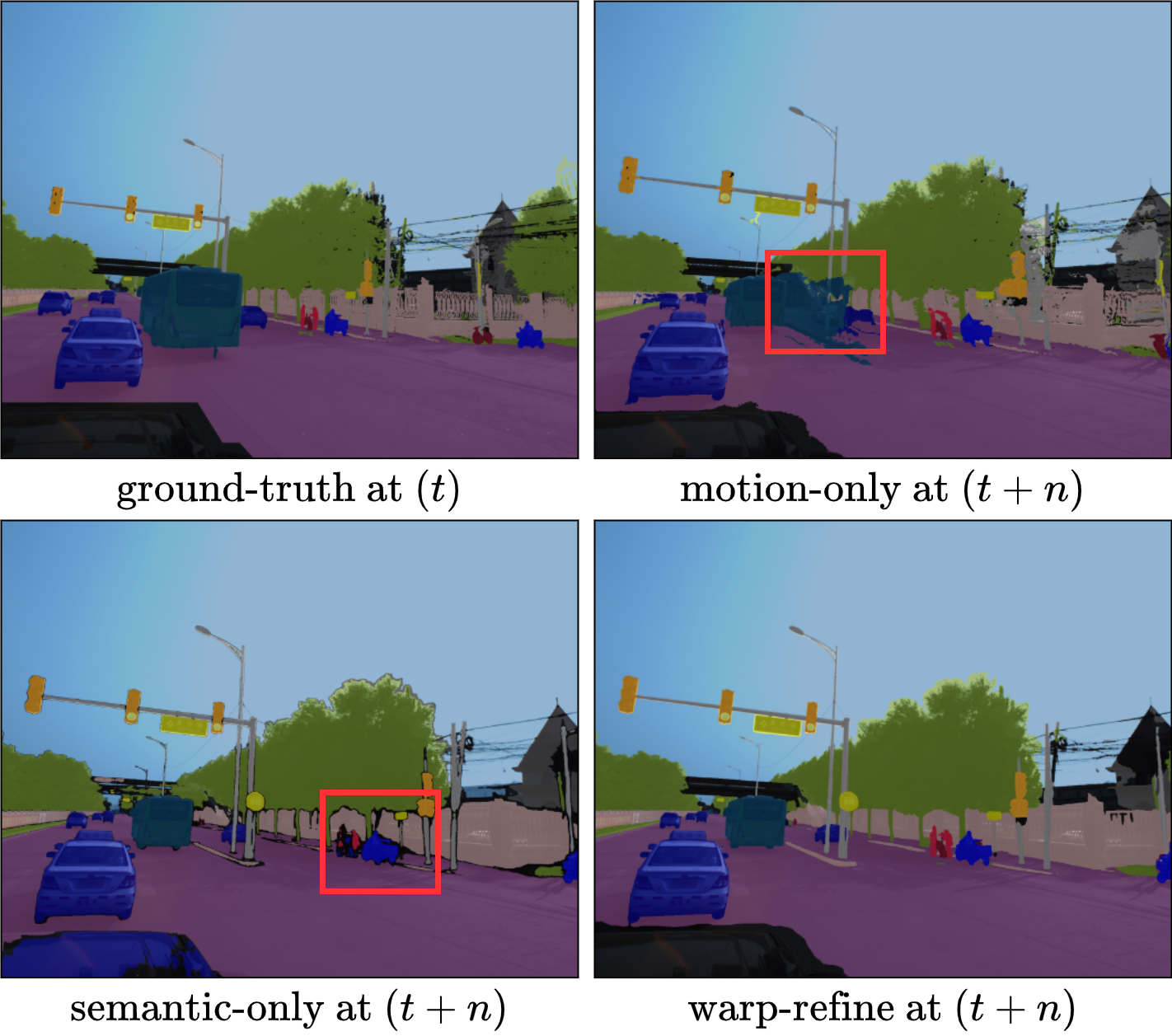}
	\caption{\small  Qualitative comparison: We compare our method  \emph{warp-refine} against other auto-labeling methods  on ApolloScape~\cite{as_dataset}. As our method combines semantic and motion cues, mistakes in one (e.g., caused by rare class, or new scene elements) are corrected by the other, leading to superior results.}
	\label{fig_as_qualitative}
\end{figure}

%% file: tables/nyu_ablation.tex
    
\begin{table*}[t]
    \centering
    \caption{\small We analyse the performance of semantic-segmentation models trained with auto-labelled data. Across the three datasets, we observe that \textit{warp-refine} labels consistently induce larger improvements than labels from other methods. 
    The results for \textit{semantic-only}, \textit{warp-inpaint} and \textit{warp-refine} are computed by sampling auto-labelled frames at time steps $(t\pm \{2,4,6,8\})$. Following Zhu et al.~\cite{nvidia_cvpr19}, for \textit{motion-only}, we only sample frames at time steps $(t\pm 2)$. We report the average performance (\small $\mu(\textit{mIoU})$), sample standard deviation (\small  $\sigma(\textit{mIoU})$) and average improvement over baseline ($\Delta(\textit{mIoU})$) by performing three runs of each experiment.
    }
    \label{table:nyu_ablation}
    \begin{tabular}{llccccccccc}
\toprule
\multicolumn{2}{l}{Dataset}  & \multicolumn{3}{c}{ NYU-V2~\cite{nyu_dataset} } & \multicolumn{3}{c}{ ApolloScape~\cite{as_dataset} } &  \multicolumn{3}{c}{ Cityscapes val-split~\cite{cs_dataset} } \\
\multicolumn{2}{c}{} &{\small $\mu(\textit{mIoU})$} &{\small  $\sigma(\textit{mIoU})$ }&{\small $ \Delta(\textit{mIoU})$ }&{\small  $\mu(\textit{mIoU})$ }&{\small $\sigma(\textit{mIoU})$ }&{\small $\Delta(\textit{mIoU})$ }&{\small $\mu(\textit{mIoU})$ }&{\small $\sigma(\textit{mIoU})$ }&{\small $\Delta(\textit{mIoU})$ } \\
\midrule
\multicolumn{2}{l}{\small Baseline}  & $50.58$ & $0.25$& - & $72.63$ & $0.09$& - & $83.35$ & $0.10$ & - \\
\multicolumn{2}{l}{\textit{\small motion-only}}& $50.38$ & $0.13$  & $-0.20$ & $72.57$ & $0.15$ & $-0.16$ & $83.01$ & $0.13$ & $-0.34$\\
\multicolumn{2}{l}{\textit{\small semantic-only}} & $50.93$  & $0.17$ & $+0.35$ & $73.10$& $0.14$ & $+0.47$ & $83.91$ & $0.04$ & $+0.56$\\
\midrule
\multicolumn{2}{l}{\textit{\small warp-inpaint}}  & $51.28$   & $0.19$ & $+0.70$ & $ 72.97$ & $ 0.22$ & $ +0.34 $ & - & - & -\\
\multicolumn{2}{l}{\textit{\small warp-refine}} & $\mathbf{52.12}$  & 0.13 & $\mathbf{+1.54}$ & $ \mathbf{73.74}$ & $ 0.12$ & $ \mathbf{+1.11}$ & $\mathbf{84.07}$ & $0.08$ & $\mathbf{+0.71}$\\
\midrule
\multicolumn{2}{l}{\small Oracle (GT)} & - & - & - & $75.36$ & $0.07$ & $+2.73$ & - & - & - \\
\bottomrule
    \end{tabular}
\end{table*}

%% file: figures/figure_qualitative_cs.tex
\begin{figure}[t]
	\centering
	\includegraphics[width=1.00\linewidth]{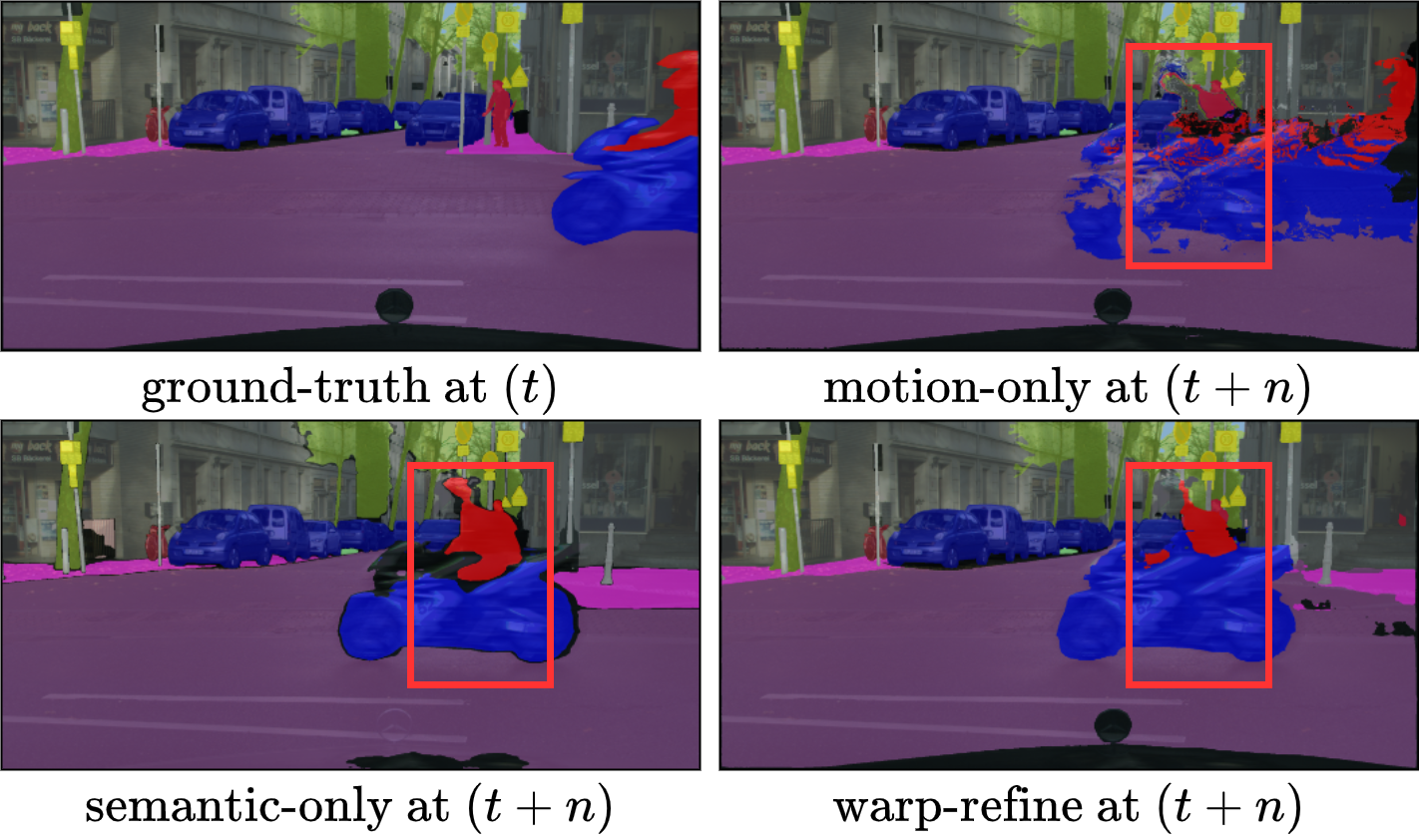}
	\caption{\small  Qualitative comparison: We show a failure case for our method \emph{warp-refine} on Cityscapes~\cite{cs_dataset}. Our method can fail when both semantic and motion cues fail (e.g. `Rider' is mislabeled). }
	\label{fig_cs_qualitative}
\vspace{-0.5em}
\end{figure}

%% file: figures/figure_ablation_1.tex
\begin{figure}[t!]
	\centering
	\includegraphics[width=1.0\linewidth]{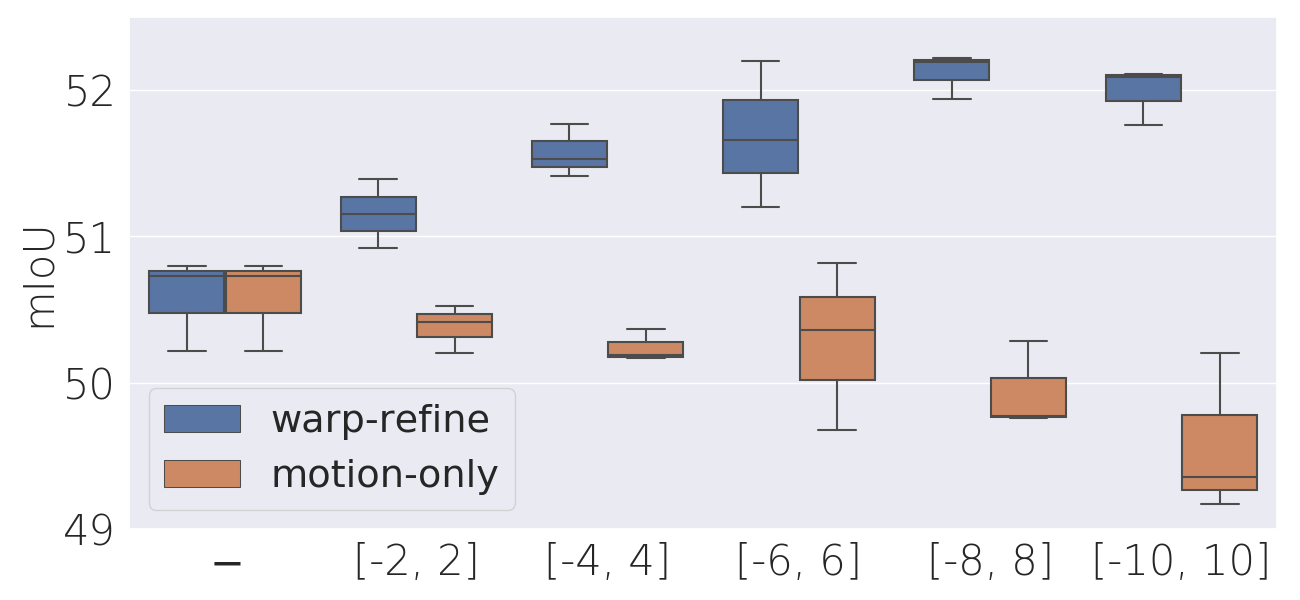}
	\caption{\small We evaluate the performance of semantic-segmentation model trained with auto-labeled frames sampled in different time-intervals. 
	Due to the accumulation of errors in \textit{motion-only} labels, performance drops as we sample frames further away. In contrast, \emph{warp-refine} labels are especially useful at larger time-steps ($\pm10$). This ablation is performed on the NYU-V2 dataset~\cite{nyu_dataset}.}
	\label{fig_nyu_seq_size}
\vspace{-1.0em}
\end{figure}

%% file: tables/table_sota_v2.tex
\begin{table*}[t]
  \caption{\small Comparison against \textit{state-of-the-art} methods on the Cityscapes~\cite{cs_dataset} test-split. All the methods reported here are trained with extra data (i.e. pretraining on Mapillary dataset~\cite{mapillary} and using coarse-labels). Our method is comparable to the current \textit{state-of-the-art} models on Cityscapes, attaining $0.1$ \textit{mIoU} more than the previous state-of-the-art method (Chen et al.~\cite{google_student}). 
}
  \label{tab_cs_sota_v2}
  \centering
  \vspace*{-1.5ex}
  \scalebox{0.73}{
  \begin{tabular}{lccccccccccccccccccccc}
    \toprule
     \ & \rotatebox{90}{road}& \rotatebox{90}{sidewalk}& \rotatebox{90}{building} &\rotatebox{90}{wall} &\rotatebox{90}{fence} &\rotatebox{90}{pole} &\rotatebox{90}{T. light} & \rotatebox{90}{T. sign} &\rotatebox{90}{vege.} &\rotatebox{90}{terrain} &\rotatebox{90}{sky} &\rotatebox{90}{person} &\rotatebox{90}{rider} &\rotatebox{90}{car} &\rotatebox{90}{truck} &\rotatebox{90}{bus} &\rotatebox{90}{train} &\rotatebox{90}{M. bike} &\rotatebox{90}{bike} &mIoU \\
    \midrule
				
				Zhu et al.~\cite{nvidia_cvpr19}    &  $98.8$     &     $87.8$   &  $94.2$     &     $64.1$ &  $65.0$     &     $72.4$ &  $79.0$     &     $82.8$ &  $94.2$     &     $74.0$ &  $96.1$     &     $88.2$ &  $75.4$     &     $96.5$ &  $78.8$     &     $94.0$ &  $91.6$     &     $73.8$ &  $79.0$     &     $83.5$     \\
				
				Yuan et al.~\cite{ocr_eccv_20}   &  $98.9$     &     $88.3$   &  $94.4$     &     $68.0$ &  $67.8$     &     $73.6$ &  $ 80.6$     &     $83.9$ &  $94.4$     &     $74.4$ &  $96.0$     &     $89.2$ &  $75.8$     &     $96.8$ &  $83.6$     &     $94.2$ &  $91.3$     &     $74.0$ &  $80.1$     &     $84.5$       \\
				Tao et al.~\cite{tao2020hierarchical}       & $99.0$ & $89.2$ & $94.9$ & $71.6$ & $69.1$ & $75.8$ & $82.0$ & $85.2$ & $94.5$ & $75.0$ & $96.3$ & $90.0$ & $79.4$ & $96.9$ & $79.8$ & $94.0$ & $85.8$ & $77.4$ & $81.4$ & $85.1$ \\
				Chen et al.~\cite{google_student}    & $98.8$ & $88.3$ & $94.6$ & $65.3$ & $69.6$ & $75.2$ & $80.9$ &  $84.4$ & $94.3$ & $74.4$ & $96.2$ & $90.0$ & $79.7$ & $96.7$ & $83.0$ & $95.6$ & $93.4$ & $78.4$ & $79.6$  &  $85.2$ \\
				\midrule
				  Ours   &  $98.9$     &     $88.6$   &  $94.9$     &     $71.5$ &  $68.7$     &     $75.5$ &  $82.0$     &     $85.2$ &  $94.5$     &     $74.1$ &  $96.3$     &     $89.9$ &  $79.4$     &     $96.9$ &  $80.2$     &     $94.4$ &  $92.5$     &     $75.7$ &  $81.2$     &     $\mathbf{85.3}$        \\
				
				\bottomrule
 \end{tabular}}
\vspace{-0.5em}
\end{table*}

%% file: tables/nyu_sota.tex
\begin{table}[t]
    \label{tab_apolloscape}

    \begin{tabular}{lcccccc}
    \toprule
    Method & $\textit{pixel-acc}$ & $\textit{mean-acc}$ & $\textit{mIoU}$ \\
    \midrule
    TRL-ResNet50 \cite{Zhang_2018_ECCV} & $76.2$ & $56.3$ & $46.4$ \\
    MTI Net~\cite{vandenhende2020mtinet} & $75.3$ & $62.9$ & $49.0$\\
    PAD-Net \cite{Xu_2018_CVPR} & $75.2$ & $62.3$ & $50.2$ \\
    PAP-Net \cite{Zhang_2019_CVPR} & $76.2$ & $62.5$ & $50.4$ \\
    \midrule
    Ours & $\mathbf{77.6}$ & $\mathbf{66.0}$ & $\mathbf{52.2}$\\
    \bottomrule
    \end{tabular}
    \centering
    \caption{\small Comparison against \textit{state-of-the-art} methods on NYU-V2~\cite{nyu_dataset} test-split: Our model trained with \emph{warp-refine} labels attains a large jump of $1.8$ in terms of \textit{mIoU}.
    }
    \label{tab_nyu_sota}
\vspace{-0.5cm}
\end{table}

%% file: tables/kitti_sota.tex
\begin{table}[t]
	\begin{center}
			\begin{tabular}{lcccc}
				\toprule
				& \multicolumn{2}{c}{Class} & \multicolumn{2}{c}{Category} \\
				Method	&  $\textit{mIoU}$  &	$\textit{iIoU}$  & $\textit{mIoU}$ &	$\textit{iIoU}$   \\
				\toprule
				SegStereo \cite{Yang_2018_ECCV}   & $59.10$	& $28.00$	& $81.31$	& $60.26$ \\
				AHiSS \cite{meletis_iv18}	 & $61.24$	& $26.94$	& $81.54$	& $53.42$ \\
				LDN2 \cite{Krapac_iccv}	 & $63.51$	& $28.31$	& $85.34$	& $59.07$ \\
				MapillaryAI \cite{Bulo_2018_CVPR}	 & $69.56$	& $43.17$	& $86.52$	& $68.89$  \\
				Zhu et al.~\cite{nvidia_cvpr19} 	 & $72.83$ & $48.68$ & $88.99$  & \textbf{75.26}    \\
				\hline
				Ours & \textbf{76.44}	& \textbf{50.92}	& \textbf{89.63} & 73.69 \\
				\bottomrule
			\end{tabular}
		\caption{\small Comparison against \textit{state-of-the-art} methods on the KITTI~\cite{kitti_dataset} test-split: Our model trained with \emph{warp-refine} labels improves over the prior work by a notable margin of $3.61$ $\textit{mIoU}$.}
		\label{tab_kitti_sota}
		\vspace{-4ex}
	\end{center}
\end{table}

%% file: sub_2/conclusions.tex
In this work, we propose a novel approach for video auto-labeling: \textit{Warp-Refine Propagation}, which combines geometric and semantic cues for label propagation. By leveraging the concept of \textit{cycle-consistency} across time, our method learns to \textit{refine} propagated labels in a \textit{semi-supervised} setting. With \textit{warp-refine}, we can accurately propagate labels for long time horizons (i.e. $\pm 10$ frames). Via a detailed ablative analysis we show that \textit{warp-refine} surpasses previous auto-labeling methods by a notable margin of $13.1$ \textit{mIoU} on \textit{ApolloScape}. Further, labels generated from \textit{warp-refine} are shown to be useful for improving single-frame semantic-segmentation models. By training semantic segmentation models with \textit{warp-refine} labels, we achieve state-of-the-art performance on \textit{NYU-V2} ($+1.8$ \textit{mIoU}), \textit{KITTI} ($+ 3.6$ \textit{mIoU}) and \textit{Cityscapes} ($+ 0.1$ \textit{mIoU}). The optimal way to combine the propagated labels with manual annotations and weaker sources of supervision (i.e. coarse labels) remains an unsolved problem, which we aim to address in our future work. 

\vfill

\noindent
\textbf{Acknowledgments.}~This work was supported by Woven Core, Inc.

%% file: subsections/supp.tex
\section{Supplementary material}
\label{section-suppl}
In this appendix, we provide additional details and experimental results to help further understand and reproduce our proposed method, i.e. \emph{Warp-Refine Propagation}.
The supplementary material is divided into the following sections: 

\begin{itemize}
    \item \textbf{Section~\ref{additional_exp} - Additional experimental studies}: We explore different aspects of training with propagated labels, such as training with different architectures, and training in data scarce setting.
    \item \textbf{Section~\ref{section_details} - Additional details}: We provide details for our training setup, as well as details of ApolloScape dataset~\cite{as_dataset} usage. 
    \item \textbf{Section~\ref{section_qualitative} - Qualitative results}: We provide qualitative examples visualizing the results of different aspects of our method. 
\end{itemize}

\section{Additional experimental studies}
\label{additional_exp}

\let\clearpage\relax

\input{tables/table_main_ablation_cityscapes}

\subsection{Propagation for different architectures}
We evaluate the benefit of the propagated labels on different semantic segmentation models. This result is summarized in Table~\ref{tab_main_ablation}. We see that the propagated labels are significantly beneficial for smaller architectures, which have lower performance. However, in the case of motion-only propagated labels, we see that the performance is unaffected or sometimes deteriorated. Note that these results do not use the 20000 additional coarse labels, nor pretraining on Mapillary Vistas~\cite{mapillary}.

\subsection{Motion estimation model ablation}
A simple way to use geometric cues is to simply warp the labels between consecutive frames based on the Optical flow. We tried different optical flow methods including RAFT~\cite{raft_of} for warping, but found them to be unsuitable. Apart from drifting errors, directly warping with optical flow also causes content duplication
on de-occluded regions~\cite{zhao2020maskflownet}. Therefore, for warping labels between consecutive frames, we found video prediction to work the best for us.

\let\clearpage\relax

\input{sup_tex/table_zhu_1}

\subsection{Ablative analysis with motion-only labels}
\label{section_zhu}

As indicated in the main paper, we do not train using the Relaxed Label Loss (RLL) proposed by Zhu et al.~\cite{nvidia_cvpr19}, and also use a fixed epoch-size. In this section, we provide additional ablative experiments, validating our choices. Our results are summarized in Table~\ref{table_zhu}, along with the numbers reported by Zhu et al.~\cite{nvidia_cvpr19} under similar training conditions. Note that we add the \textit{motion-only} propagated labels at time step $t \pm 3$ as represented by $D_3^m$. We perform the experiments under different training settings, namely considering the usage of coarse-labels and Mapillary Vistas pre-training. We report the mean and the standard deviation by conducting three runs with different random number generator seeds for each result. We note a significant improvement between our baseline when training with the Cityscapes coarse-labels and with Mapillary Vistas pre-training ($80.94$ \textit{mIoU}) and the baseline reported in~\cite{nvidia_cvpr19} ($79.46$ \textit{mIoU}) which we attribute to a longer training schedule of our baseline and a modified learning rate schedule:

\begin{enumerate}
    \item \emph{Longer training of the baseline}:~When training with propagated labels, our dataset size for $\mathtt{B} + {D}_3^m$ is increased. This leads to more training iterations for $\mathtt{B} + {D}_3^m$ with respect to the baseline model ($\mathtt{B}$ is trained for only one-third the iterations of $\mathtt{B} + {D}_3^m$). We therefore modify the training such that $\mathtt{B}$ is trained for the same number of iterations as $\mathtt{B} + {D}_3^m$.
    \item \emph{Higher learning rate for the baseline}:~We increase the learning rate by a factor of $8$ for the baseline $\mathtt{B}$ (we observed that the scale of cross-entropy loss is much smaller than the scale of Relaxed Label Loss).
\end{enumerate}

With the updated baseline, we find RLL as well as training with \textit{motion-only} labels to be ineffective. Further, to avoid the pitfall of under-training the baseline, we fix the epoch-size for all the models we compare. This ensure that the improvement by using additional labels is not conflated with improvement by longer training.

\section{Additional details}
\label{section_details}

\subsection{Training details}
\label{subsec_training_details}
We use an SGD optimizer and employ a polynomial learning rate policy, where the initial learning rate is multiplied by $(1 - \frac{\texttt{epoch}}{\texttt{max epoch}})^{\texttt{power}}$. The learning rate is varied for different datasets: for KITTI~\cite{kitti_dataset} we utilize a learning rate of $0.0005$, for Cityscapes we utilize $0.01$ and for NYU-V2~\cite{nyu_dataset} we utilize $0.001$. Momentum and weight decay are set to $0.9$ and $0.0001$ respectively. We use synchronized batch normalization (batch statistics synchronized across all GPUs) with the batch distributed over 8 V100 GPUs. For data augmentation, we randomly scale the input images (from $0.5$ to $2.0$), and apply horizontal flipping, Gaussian blur and color jittering during training. Further, we utilize uniform sampling~\cite{nvidia_cvpr19} across semantic classes with $50\%$ of each epoch.

We introduce two changes from the training configuration outlined by Zhu et al.~\cite{nvidia_cvpr19}:
\begin{itemize}
    \item As our approach generates additional training data, the epoch size varies greatly depending on training settings. This can lead to a situation where the observed improvement in performance can be due to longer training rather than generated data (As shown in Section~\ref{section_zhu}). To avoid such mis-attribution of the reason for improvement, we ensure that the training regime for all compared experiments is equivalent. To achieve that, we define an epoch to have a fixed size (roughly $3 \times$ the size of the normal dataset). With this definition, we train for 175 epochs. %
    \item We adjust our data sampling such that in each epoch, $30\%$ samples are drawn from the manually annotated dataset, and $70\%$ data is drawn from the generated dataset (through label propagation). Hence, the number of pseudo-labels considered per epoch remains consistent independent of the amount of generated labels (In the presence of Coarse labelled data, we reduce sampling from the generated dataset to 30\%). %
\end{itemize}

For models evaluated on the test set, we use the same training validation split used by Zhu et al.~\cite{nvidia_cvpr19} ($\mathtt{cv2}$ split). The cities Mönchengladbach, Strasbourg and Stuttgart are used as validation set while all the other cities are used as training data.

\subsection{ApolloScape partitioning}
The ApolloScape dataset~\cite{as_dataset} contains pixel-level annotations for sequentially recorded images, divided as 40960 training and 8327 validation images. These images are further broken into the subsets based on the road on which they were recorded, and the Record-ID. Each Record-ID consists of variable length sequentially annotated frames. We break these sequentially annotated frames into partitions each consisting of 21 consecutive frame. The images which are not a part of any such 21-frame partition (for example when a Record-ID contains less than 21 frames) are discarded.
 
Now, from each partition, we utilize the central frame as a training data point (i.e. with manual annotation) and all the other frames are treated as frames where labels have to be generated via propagation. This allows us to create a dataset with ground-truth labels containing $2005$ frames, and additional $40100$ sequential images (we only use the provided ground truth for these images for evaluation purposes).
 
\let\clearpage\relax
\input{sup_tex/fig_cyclic}

Note that to ensure that training and validation data do not have any overlap (which could happen if any partition of 21 frames contains validation samples), we combine the training and validation subset, and re-divide it at a Record-ID level (randomly). This ensures that none of our train-sequences have any overlap with the validation data. Due to this our training and validation split are different from the one provided with the dataset. To encourage and facilitate comparisons with our work, we will release our training and validation splits to the community.

\let\clearpage\relax
\input{sup_tex/fig_denoiser}

\subsection{Denoising module}

Our denoising module $\Omega_\lambda$ is inspired from semantic-to-real models~\cite{pix2pix, spades}. We show our architecture in Figure~\ref{fig_denoiser}. Our network takes the warp-inpainted labels $L^w_t$, along with auxiliary inputs: the warped image $I^m_t$, and the image at time $t+1$ $I_{t+1}$ to generate refined labels $L^R_{t+1}$:
\begin{align}
    L^R_{t+1} = \Omega_\lambda(I_{t+1}, I^w_{t}, L^w_{t})
\end{align}

The warped labels $l^w_t$ are used as one-hot vectors per pixel. All the inputs are concatenated along the ``channel'' dimension and provided to the encoder network $N_{encoder}$. The generated encoding is then concatenated with OCR-features~\cite{ocr_eccv_20} of the image $I_{t+1}$ (extracted using the baseline model $g_\psi$ trained with only manually annotated images). This is done to provide rich semantic cues for regions with new objects. Finally the concatenated encoding is passed through the decoder network $N_{decoder}$ to generate the refined labels $L^R_{t+1}$. The complete pipeline is visualized in Figure~\ref{fig_denoiser}.

Our network is trained with the same optimization setting as detailed in Section~\ref{subsec_training_details}. The RMI loss~\cite{rmi_loss} is used to compute the cycle-consistency loss $\mathcal{L}(L_t, L^R_t)$.

\section{Qualitative results}
\label{section_qualitative}

\let\clearpage\relax
\input{sup_tex/fig_kitti_examples}

In Figure~\ref{fig_cyclic}, we show examples of cyclic warped labels $l^{\circ}$ (cf. Section 3.2 in the main paper) for different cycle lengths. As shown, by using different cycle lengths we are able to expose the denoising module $\Omega_\lambda$ to a larger variety of label noise created due to warp-inpaint propagation. 
\noindent
Figure~\ref{fig_kitti_examples} compares the output of model trained with and without \textit{warp-refine} labels on KITTI~\cite{kitti_dataset} test-split (and nearby images using scene-flow test-split). We observe that on training with \textit{warp-refine} labels, improves the networks performance on confusing classes such as (i) bus-truck, (ii) truck-car, (iii) rider-pedestrian, and (iv) fence-wall. 

\let\clearpage\relax
\input{sup_tex/fig_cs_examples}

\noindent
Finally, Figure~\ref{fig_cs_examples} shows additional qualitative comparisons between our propagation method and established baselines: i) \textit{motion-only} labels~\cite{nvidia_cvpr19}, and ii) \textit{semantic-only} labels~\cite{tao2020hierarchical}. (a)-(d) show cases where our approach surpasses the other methods significantly. We also highlight the errors we observe in our method: 1) Our labels are weak for fine edges, 2) Our labels still appear to show some warping noise (as shown in example (f)) and 3) Our labels can sometimes mislabel some classes (as shown in example (e)). Note that examples in Figure~\ref{fig_cs_examples} are generated with DeepLabv3 (ResNeXt-50)~\cite{deep_v3, resnext} architecture for $g_\psi$.

%% file: tables/table_main_ablation_cityscapes.tex
\begin{table*}[t]
    \centering
    \caption{\small  Training with different labelling policies on Cityscapes~\cite{cs_dataset} val-split: We evaluate the benefit from warp-refine propagation across different segmentation models. Due to the lack of semantic complexity in the dataset (only 19 classes), and the high performance (\textit{mIoU} $= 83.35$) of the semantic labelling network, we find the \textit{semantic-only} labels to give significant benefits as well. (Note that for \textit{motion-only} we utilize only time-frames $\pm[2]$ as recommended by the authors Zhu et al.~\cite{nvidia_cvpr19}). We report the average of three independent runs with different random seeds. Note that we do not use any additional data (coarse labels and Mapillary Vistas~\cite{mapillary} pretraining) for this ablative analysis.
    }
\vspace{-1.5ex}
    \label{tab_main_ablation}
    \begin{tabular}{llccccc}
\toprule
\multicolumn{2}{l}{Model}& Backbone & \multicolumn{1}{c}{Baseline}& \multicolumn{1}{c}{\textit{motion-only}}& \multicolumn{1}{c}{\textit{semantic-only}} & \multicolumn{1}{c}{\emph{warp-refine}} \\
\midrule
\multicolumn{2}{l}{DeepLab V3~\cite{deep_v3} }& ResNeXt-50~\cite{resnext} & $79.26$ &$79.01$  & $80.45$  & $\mathbf{80.68}$  \\
\multicolumn{2}{l}{OCR Net~\cite{ocr_eccv_20} }& ResNeXt-50~\cite{resnext} & $79.55$ & $79.60$ & $\mathbf{80.89}$ & $80.80$ \\
\multicolumn{2}{l}{MSA-HRNet-OCR~\cite{tao2020hierarchical}} & HRNet-W48~\cite{sun2019highresolution} & $83.35$  & $83.00$ & $83.91$ & $\mathbf{84.07}$ \\
\bottomrule
    \end{tabular}
    
\vspace{-0.25cm}
\end{table*}

%% file: sup_tex/table_zhu_1.tex
\begin{table*}[t]
    \centering
    \caption{\small  Results of training a segmentation model with labels generated by video propagation~\cite{nvidia_cvpr19} ($D_3^m$) and relaxed label loss (RLL), on the Cityscapes validation split.  These experiments are conducted under different training settings (as shwon by the top two rows) . We also compare the mean IoU to those reported in previous work~\cite{nvidia_cvpr19}. We conduct three runs with different random seeds.
    }
    \label{table_zhu}
    \begin{tabular}{l|cc|cc|cc|c}
\toprule
\multicolumn{1}{c|}{Coarse Labels} & \multicolumn{2}{c|}{\xmark} & \multicolumn{2}{c|}{\cmark} & \multicolumn{2}{c|}{\cmark}  & {\small Training} \\
\multicolumn{1}{c|}{Map. Pre-train}& \multicolumn{2}{c|}{\xmark} & \multicolumn{2}{c|}{\xmark} & \multicolumn{2}{c|}{\cmark}  & {\small iterations} \\
\hline
\multicolumn{1}{c|}{}& avg. mIoU & std. & avg. mIoU & std. & avg. mIoU & std.&  \# \\
\hline
\multicolumn{1}{l|}{\cite{nvidia_cvpr19} baseline~$\mathtt{B}$ } & - & - & - & - & 79.46 & - & 175 $\times$ 2975 \\
\multicolumn{1}{l|}{\cite{nvidia_cvpr19}~$\mathtt{B}$ + RLL} & - & - & - & - & 80.85 & - & 175 $\times$ 2925 \\
\multicolumn{1}{l|}{\cite{nvidia_cvpr19}~$\mathtt{B}$ + RLL + $D_3^m$} & - & - & - & - & 81.35 & - & 175 $\times$ 8925 \\
\hline
\multicolumn{1}{l|}{Baseline $\mathtt{B}$} & 77.66 & 0.27 & 79.15&0.23 & 80.94 & 0.10 & 175 $\times$ 8925\\
\multicolumn{1}{l|}{$\mathtt{B}$+ RLL } & 77.50 & 0.12 & - & - & 80.76 & 0.18 & 175 $\times$ 8925\\
\multicolumn{1}{l|}{$\mathtt{B}$ + RLL + $D_3^m$} & 77.41 & 0.22 & 78.69 & 0.20 & 80.8 &  0.11 & 175 $\times$ 8925 \\
\bottomrule
    \end{tabular}
\end{table*}

%% file: sup_tex/fig_cyclic.tex
\begin{figure*}[t]
	\centering
	\includegraphics[width=0.97\linewidth]{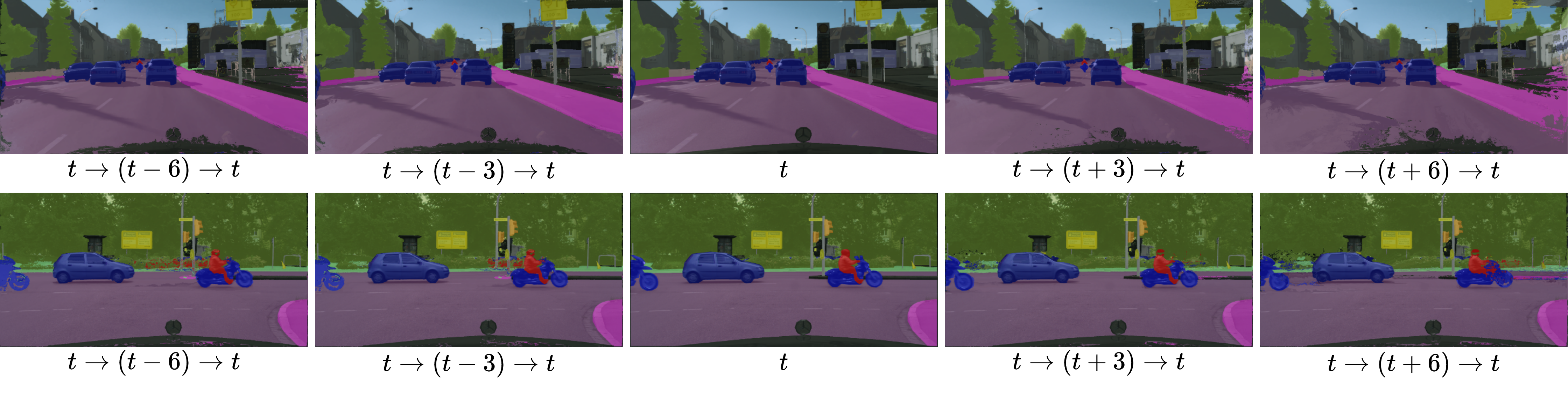}
    \vspace{-1.0em}
	\caption{\small We show examples of cyclic warped labels, generated to train the denoising network $\Omega_\lambda$. The network is trained to map the samples ($t \rightarrow t+p \rightarrow t$) to the ground truth label ($t$). Using longer cycle of propagation (higher $p$) allows us to expose the network $\Omega_\lambda$ to larger amount of warping noise.}
	\label{fig_cyclic}
\end{figure*}

%% file: sup_tex/fig_denoiser.tex
\begin{figure}[t]
	\centering
	\includegraphics[width=0.9\linewidth]{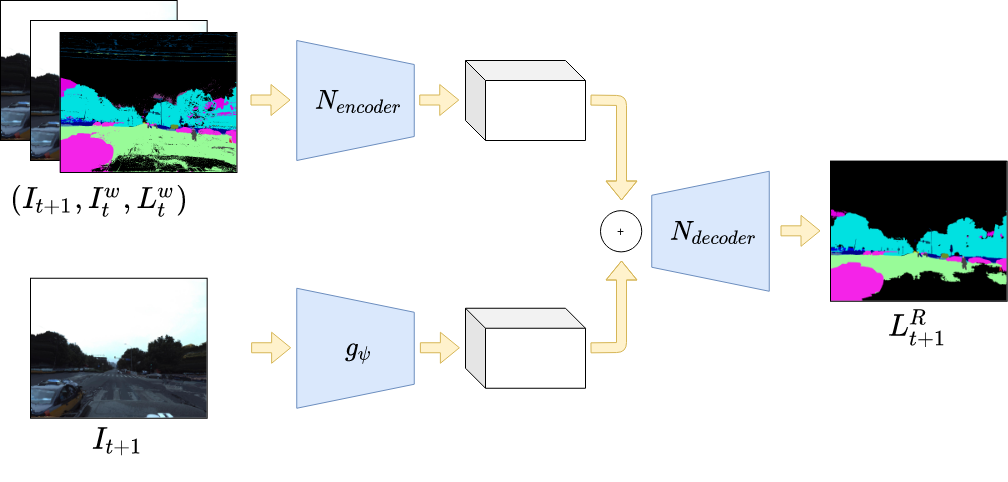}
    \vspace{-1.0em}
	\caption{Architecture of the denoiser: The encoder and decoder are based on pix2pix~\cite{pix2pix}. $g_{\psi}$ is the baseline model trained only with manually annotated labels. The input to the encoder are concatenated along the channels dimension. Similarly, the input of the decoder is the concatenated output of the encoder, and OCR-features~\cite{ocr_eccv_20} from $g_{\psi}$.}
	\label{fig_denoiser}
\end{figure}

%% file: sup_tex/fig_kitti_examples.tex
\begin{figure*}
	\centering
	\includegraphics[width=1.0\linewidth]{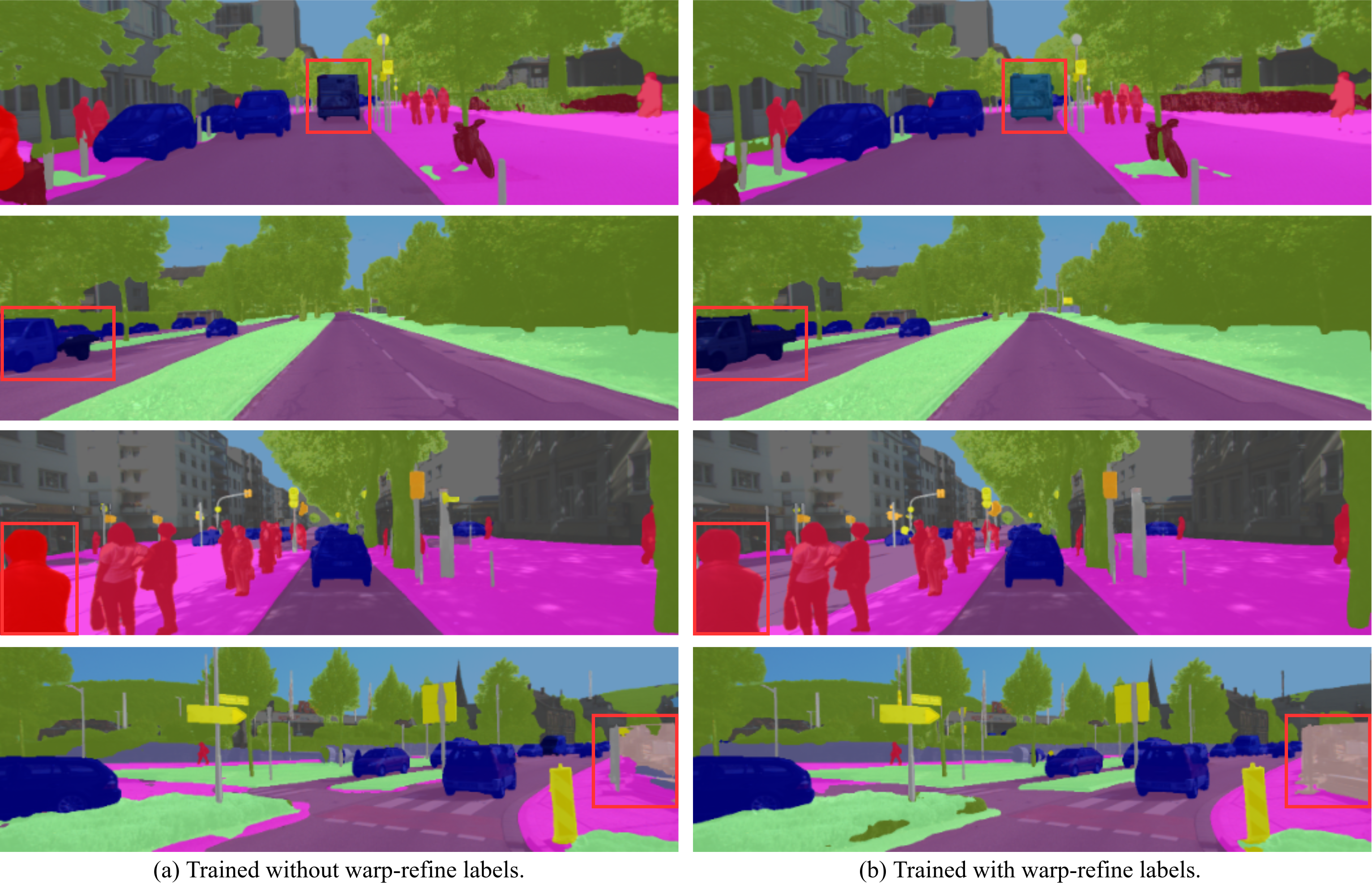}
    \vspace{-1.0em}
	\caption{\small Qualitative comparison of model trained with and without warp-refine labels. We see that training with warp-refine labels increase performance for confusing classes: Baseline model mis-predicts (i)  'bus' as 'truck', (ii) 'truck' as 'car', (iii) 'pedestrian' as 'rider', and (iv) 'fence' as 'wall' and 'sidewalk'.}
	\label{fig_kitti_examples}
\end{figure*}

%% file: sup_tex/fig_cs_examples.tex
\begin{figure*}[t]
	\centering
	\includegraphics[width=0.90\linewidth]{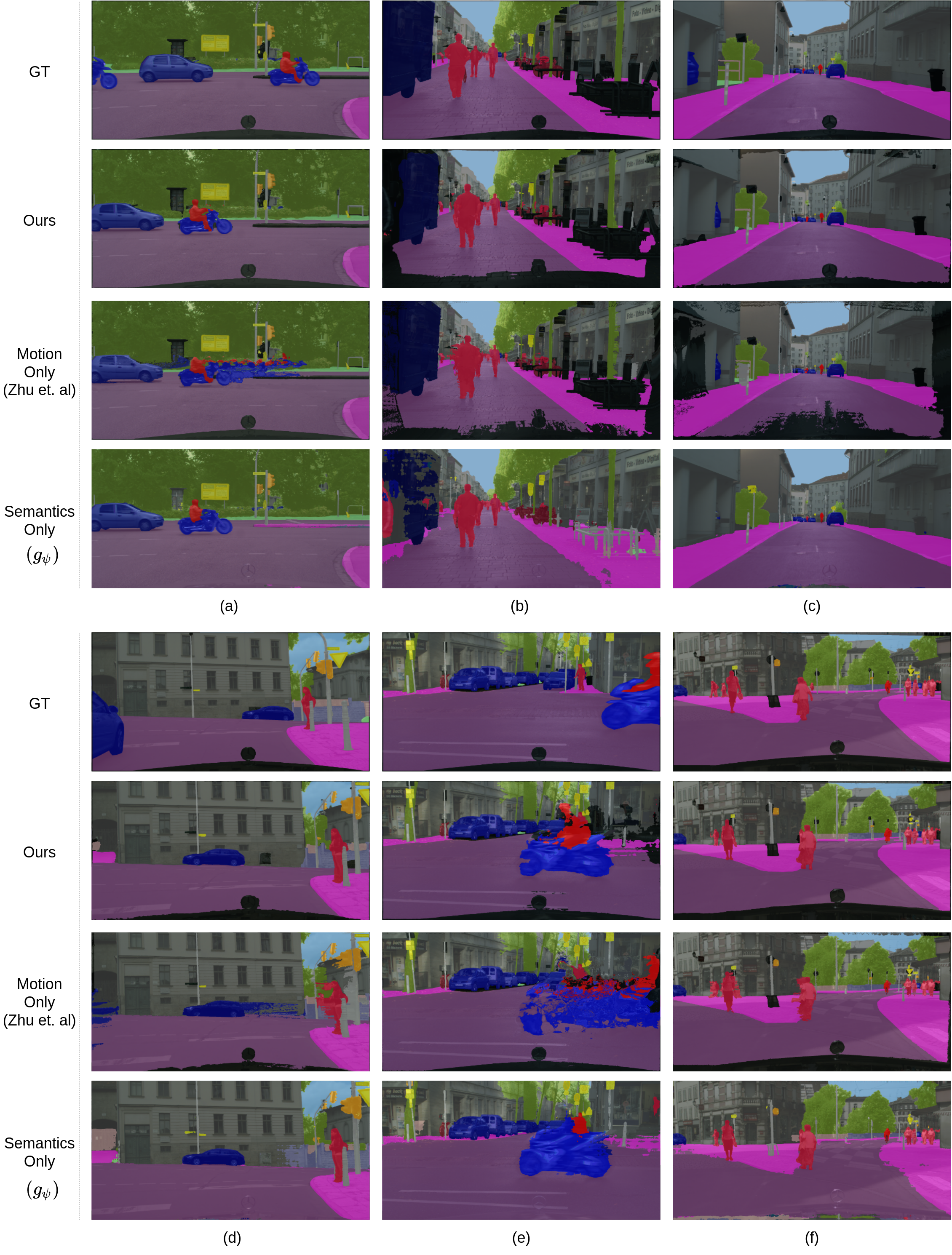}
    \vspace{-1.0em}
	\caption{\small (a) $g_\psi$ mislabels the Rider's legs, and motion only (Zhu et al.~\cite{nvidia_cvpr19}) shows heavy drifting. (b) $g_\psi$ mislabels the truck, and Zhu et al.~\cite{nvidia_cvpr19} cause drifting near the pedestrian pixels. (c) $g_\psi$ mislabels thin objects like poles (left side) (d) $g_\psi$ mislabels part of the building. (e) We note that when both semantic and motion cues fail, our method fails as well. (f) Our method outputs slightly warped labels when the consecutive frames do not contain any ego-motion (note the warping of the pole in the center).  }
	\label{fig_cs_examples}
\end{figure*}

%% file: main.bbl
\begin{thebibliography}{10}\itemsep=-1pt

\bibitem{kitti_dataset}
Hassan Alhaija, Siva Mustikovela, Lars Mescheder, Andreas Geiger, and Carsten
  Rother.
\newblock Augmented reality meets computer vision: Efficient data generation
  for urban driving scenes.
\newblock {\em International Journal of Computer Vision (IJCV)}, 2018.

\bibitem{lp_2013}
Vijay Badrinarayanan, Ignas Budvytis, and Roberto Cipolla.
\newblock Semi-supervised video segmentation using tree structured graphical
  models.
\newblock {\em IEEE Transactions on Pattern Analysis and Machine Intelligence},
  35(11):2751--2764, 2013.

\bibitem{lp_2010}
V. {Badrinarayanan}, F. {Galasso}, and R. {Cipolla}.
\newblock Label propagation in video sequences.
\newblock In {\em 2010 IEEE Computer Society Conference on Computer Vision and
  Pattern Recognition}, pages 3265--3272, 2010.

\bibitem{lp_iccvw}
I. Budvytis, P. Sauer, T. Roddick, K. Breen, and R. Cipolla.
\newblock Large scale labelled video data augmentation for semantic
  segmentation in driving scenarios.
\newblock In {\em 5th Workshop on Computer Vision for Road Scene Understanding
  and Autonomous Driving in IEEE International Conference on Computer Vision
  (ICCV)}, October 2017.

\bibitem{Bulo_2018_CVPR}
Samuel~Rota Bulò, Lorenzo Porzi, and Peter Kontschieder.
\newblock In-place activated batchnorm for memory-optimized training of dnns.
\newblock In {\em Proceedings of the IEEE Conference on Computer Vision and
  Pattern Recognition (CVPR)}, June 2018.

\bibitem{domain_seg_9}
Chaoqi Chen, Weiping Xie, Wenbing Huang, Yu Rong, Xinghao Ding, Yue Huang,
  Tingyang Xu, and Junzhou Huang.
\newblock Progressive feature alignment for unsupervised domain adaptation.
\newblock In {\em The IEEE Conference on Computer Vision and Pattern
  Recognition (CVPR)}, 2019.

\bibitem{deeplab_v1}
Liang{-}Chieh Chen, George Papandreou, Iasonas Kokkinos, Kevin Murphy, and
  Alan~L. Yuille.
\newblock Semantic image segmentation with deep convolutional nets and fully
  connected crfs.
\newblock In Yoshua Bengio and Yann LeCun, editors, {\em 3rd International
  Conference on Learning Representations, {ICLR} 2015, San Diego, CA, USA, May
  7-9, 2015, Conference Track Proceedings}, 2015.

\bibitem{deeplab_v2}
Liang{-}Chieh Chen, George Papandreou, Florian Schroff, and Hartwig Adam.
\newblock Rethinking atrous convolution for semantic image segmentation.
\newblock {\em CoRR}, abs/1706.05587, 2017.

\bibitem{google_student}
Liang-Chieh Chen, Raphael~Gontijo Lopes, Bowen Cheng, Maxwell~D. Collins,
  Ekin~D. Cubuk, Barret Zoph, Hartwig Adam, and Jonathon Shlens.
\newblock Naive-student: Leveraging semi-supervised learning in video sequences
  for urban scene segmentation, 2020.

\bibitem{deep_v3}
Liang-Chieh Chen, Yukun Zhu, George Papandreou, Florian Schroff, and Hartwig
  Adam.
\newblock Encoder-decoder with atrous separable convolution for semantic image
  segmentation.
\newblock In {\em The European Conference on Computer Vision (ECCV)}, September
  2018.

\bibitem{mscoco}
Xinlei Chen, Tsung-Yi~Lin Hao~Fang, Ramakrishna Vedantam, Saurabh Gupta, Piotr
  Dollár, and C.~Lawrence Zitnick.
\newblock Microsoft coco captions: Data collection and evaluation server.
\newblock {\em arXiv preprint arXiv:1504.00325}, 2015.

\bibitem{cs_dataset}
Marius Cordts, Mohamed Omran, Sebastian Ramos, Timo Rehfeld, Markus Enzweiler,
  Rodrigo Benenson, Uwe Franke, Stefan Roth, and Bernt Schiele.
\newblock The cityscapes dataset for semantic urban scene understanding.
\newblock In {\em Proc. of the IEEE Conference on Computer Vision and Pattern
  Recognition (CVPR)}, 2016.

\bibitem{sem_warp}
Raghudeep Gadde, Varun Jampani, and Peter~V. Gehler.
\newblock Semantic video cnns through representation warping.
\newblock In {\em The IEEE International Conference on Computer Vision (ICCV)},
  Oct 2017.

\bibitem{cut_paste}
Golnaz Ghiasi, Yin Cui, Aravind Srinivas, Rui Qian, Tsung-Yi Lin, Ekin~D.
  Cubuk, Quoc~V. Le, and Barret Zoph.
\newblock Simple copy-paste is a strong data augmentation method for instance
  segmentation, 2020.

\bibitem{rcnn_1}
R. {Girshick}, J. {Donahue}, T. {Darrell}, and J. {Malik}.
\newblock Rich feature hierarchies for accurate object detection and semantic
  segmentation.
\newblock In {\em 2014 IEEE Conference on Computer Vision and Pattern
  Recognition}, pages 580--587, 2014.

\bibitem{ilg2017flownet}
Eddy Ilg, Nikolaus Mayer, Tonmoy Saikia, Margret Keuper, Alexey Dosovitskiy,
  and Thomas Brox.
\newblock Flownet 2.0: Evolution of optical flow estimation with deep networks.
\newblock In {\em Proceedings of the IEEE conference on computer vision and
  pattern recognition}, pages 2462--2470, 2017.

\bibitem{isola2017image}
Phillip Isola, Jun-Yan Zhu, Tinghui Zhou, and Alexei~A Efros.
\newblock Image-to-image translation with conditional adversarial networks.
\newblock In {\em Proceedings of the IEEE conference on computer vision and
  pattern recognition}, pages 1125--1134, 2017.

\bibitem{Krapac_iccv}
J. {Krapac} and I.~K.~S. {Šegvic}.
\newblock Ladder-style densenets for semantic segmentation of large natural
  images.
\newblock In {\em 2017 IEEE International Conference on Computer Vision
  Workshops (ICCVW)}, pages 238--245, 2017.

\bibitem{domain_seg_2}
Yunsheng Li, Lu Yuan, and Nuno Vasconcelos.
\newblock Bidirectional learning for domain adaptation of semantic
  segmentation.
\newblock In {\em The IEEE Conference on Computer Vision and Pattern
  Recognition (CVPR)}, June 2019.

\bibitem{domain_seg_10}
Qing Lian, Fengmao Lv, Lixin Duan, and Boqing Gong.
\newblock Constructing self-motivated pyramid curriculums for cross-domain
  semantic segmentation: A non-adversarial approach.
\newblock In {\em The IEEE International Conference on Computer Vision (ICCV)},
  2019.

\bibitem{fcn_cvpr}
J. {Long}, E. {Shelhamer}, and T. {Darrell}.
\newblock Fully convolutional networks for semantic segmentation.
\newblock In {\em 2015 IEEE Conference on Computer Vision and Pattern
  Recognition (CVPR)}, pages 3431--3440, 2015.

\bibitem{future_seg}
Pauline Luc, Camille Couprie, Yann LeCun, and Jakob Verbeek.
\newblock Predicting future instance segmentation by forecasting convolutional
  features.
\newblock In {\em The European Conference on Computer Vision (ECCV)}, September
  2018.

\bibitem{roi10d}
Fabian Manhardt, Wadim Kehl, and Adrien Gaidon.
\newblock {ROI-10D:} monocular lifting of 2d detection to 6d pose and metric
  shape.
\newblock In {\em {IEEE} Conference on Computer Vision and Pattern Recognition,
  {CVPR} 2019, Long Beach, CA, USA, June 16-20, 2019}, pages 2069--2078.
  Computer Vision Foundation / {IEEE}, 2019.

\bibitem{Marcu_2020_ACCV}
Alina Marcu, Vlad Licaret, Dragos Costea, and Marius Leordeanu.
\newblock Semantics through time: Semi-supervised segmentation of aerial videos
  with iterative label propagation.
\newblock In {\em Proceedings of the Asian Conference on Computer Vision
  (ACCV)}, November 2020.

\bibitem{domain_seg_7}
Ke Mei, Chuang Zhu, Jiaqi Zou, and Shanghang Zhang.
\newblock Instance adaptive self-training for unsupervised domain adaptation.
\newblock In {\em Computer Vision -- ECCV 2020}, 2020.

\bibitem{meletis_iv18}
P. {Meletis} and G. {Dubbelman}.
\newblock Training of convolutional networks on multiple heterogeneous datasets
  for street scene semantic segmentation.
\newblock In {\em 2018 IEEE Intelligent Vehicles Symposium (IV)}, pages
  1045--1050, 2018.

\bibitem{lp_eccv}
Siva~Karthik Mustikovela, Michael~Ying Yang, and Carsten Rother.
\newblock Can ground truth label propagation from video help semantic
  segmentation?
\newblock In {\em European Conference on Computer Vision}, pages 804--820.
  Springer, 2016.

\bibitem{nyu_dataset}
Pushmeet~Kohli Nathan~Silberman, Derek~Hoiem and Rob Fergus.
\newblock Indoor segmentation and support inference from rgbd images.
\newblock In {\em ECCV}, 2012.

\bibitem{mapillary}
Gerhard Neuhold, Tobias Ollmann, Samuel Rota~Bul\`o, and Peter Kontschieder.
\newblock The mapillary vistas dataset for semantic understanding of street
  scenes.
\newblock In {\em International Conference on Computer Vision (ICCV)}, 2017.

\bibitem{spades}
Taesung Park, Ming-Yu Liu, Ting-Chun Wang, and Jun-Yan Zhu.
\newblock Semantic image synthesis with spatially-adaptive normalization.
\newblock In {\em Proceedings of the IEEE Conference on Computer Vision and
  Pattern Recognition}, 2019.

\bibitem{prop_rt}
Matthieu Paul, Christoph Mayer, Luc~Van Gool, and Radu Timofte.
\newblock Efficient video semantic segmentation with labels propagation and
  refinement, 2019.

\bibitem{sdc_net}
Fitsum~A Reda, Guilin Liu, Kevin~J Shih, Robert Kirby, Jon Barker, David
  Tarjan, Andrew Tao, and Bryan Catanzaro.
\newblock Sdc-net: Video prediction using spatially-displaced convolution.
\newblock In {\em Proceedings of the European Conference on Computer Vision
  (ECCV)}, pages 718--733, 2018.

\bibitem{cycle_vid_interp}
Fitsum~A. Reda, Deqing Sun, Aysegul Dundar, Mohammad Shoeybi, Guilin Liu,
  Kevin~J. Shih, Andrew Tao, Jan Kautz, and Bryan Catanzaro.
\newblock Unsupervised video interpolation using cycle consistency.
\newblock In {\em The IEEE International Conference on Computer Vision (ICCV)},
  October 2019.

\bibitem{faster_rcnn}
Shaoqing Ren, Kaiming He, Ross Girshick, and Jian Sun.
\newblock Faster r-cnn: Towards real-time object detection with region proposal
  networks.
\newblock In C. Cortes, N.~D. Lawrence, D.~D. Lee, M. Sugiyama, and R. Garnett,
  editors, {\em Advances in Neural Information Processing Systems 28}, pages
  91--99. Curran Associates, Inc., 2015.

\bibitem{unet}
Olaf Ronneberger, Philipp Fischer, and Thomas Brox.
\newblock U-net: Convolutional networks for biomedical image segmentation.
\newblock In {\em International Conference on Medical image computing and
  computer-assisted intervention}, pages 234--241. Springer, 2015.

\bibitem{sun2019highresolution}
Ke Sun, Yang Zhao, Borui Jiang, Tianheng Cheng, Bin Xiao, Dong Liu, Yadong Mu,
  Xinggang Wang, Wenyu Liu, and Jingdong Wang.
\newblock High-resolution representations for labeling pixels and regions,
  2019.

\bibitem{tao2020hierarchical}
Andrew Tao, Karan Sapra, and Bryan Catanzaro.
\newblock Hierarchical multi-scale attention for semantic segmentation, 2020.

\bibitem{raft_of}
Zachary Teed and Jia Deng.
\newblock Raft: Recurrent all-pairs field transforms for optical flow.
\newblock In {\em European conference on computer vision}, pages 402--419.
  Springer, 2020.

\bibitem{vandenhende2020mtinet}
Simon Vandenhende, Stamatios Georgoulis, and Luc~Van Gool.
\newblock Mti-net: Multi-scale task interaction networks for multi-task
  learning, 2020.

\bibitem{feelvos2019}
Paul Voigtlaender, Yuning Chai, Florian Schroff, Hartwig Adam, Bastian Leibe,
  and Liang-Chieh Chen.
\newblock Feelvos: Fast end-to-end embedding learning for video object
  segmentation.
\newblock In {\em CVPR}, 2019.

\bibitem{domain_seg_1}
Tuan-Hung Vu, Himalaya Jain, Maxime Bucher, Matthieu Cord, and Patrick Perez.
\newblock Advent: Adversarial entropy minimization for domain adaptation in
  semantic segmentation.
\newblock In {\em The IEEE Conference on Computer Vision and Pattern
  Recognition (CVPR)}, June 2019.

\bibitem{hr_net_pami}
Jingdong Wang, Ke Sun, Tianheng Cheng, Borui Jiang, Chaorui Deng, Yang Zhao,
  Dong Liu, Yadong Mu, Mingkui Tan, Xinggang Wang, Wenyu Liu, and Bin Xiao.
\newblock Deep high-resolution representation learning for visual recognition.
\newblock {\em TPAMI}, 2019.

\bibitem{as_dataset}
Peng Wang, Xinyu Huang, Xinjing Cheng, Dingfu Zhou, Qichuan Geng, and Ruigang
  Yang.
\newblock The apolloscape open dataset for autonomous driving and its
  application.
\newblock {\em IEEE transactions on pattern analysis and machine intelligence},
  2019.

\bibitem{Wang_2020_CVPR}
Shuxin Wang, Shilei Cao, Dong Wei, Renzhen Wang, Kai Ma, Liansheng Wang, Deyu
  Meng, and Yefeng Zheng.
\newblock Lt-net: Label transfer by learning reversible voxel-wise
  correspondence for one-shot medical image segmentation.
\newblock In {\em Proceedings of the IEEE/CVF Conference on Computer Vision and
  Pattern Recognition (CVPR)}, June 2020.

\bibitem{pix2pix}
Ting-Chun Wang, Ming-Yu Liu, Jun-Yan Zhu, Andrew Tao, Jan Kautz, and Bryan
  Catanzaro.
\newblock High-resolution image synthesis and semantic manipulation with
  conditional gans.
\newblock In {\em Proceedings of the IEEE Conference on Computer Vision and
  Pattern Recognition}, 2018.

\bibitem{CVPR2019_CycleTime}
Xiaolong Wang, Allan Jabri, and Alexei~A. Efros.
\newblock Learning correspondence from the cycle-consistency of time.
\newblock In {\em CVPR}, 2019.

\bibitem{noisy_self}
Qizhe Xie, Eduard Hovy, Minh-Thang Luong, and Quoc~V Le.
\newblock Self-training with noisy student improves imagenet classification.
\newblock {\em arXiv preprint arXiv:1911.04252}, 2019.

\bibitem{resnext}
Saining Xie, Ross Girshick, Piotr Dollar, Zhuowen Tu, and Kaiming He.
\newblock Aggregated residual transformations for deep neural networks.
\newblock In {\em The IEEE Conference on Computer Vision and Pattern
  Recognition (CVPR)}, July 2017.

\bibitem{Xu_2018_CVPR}
Dan Xu, Wanli Ouyang, Xiaogang Wang, and Nicu Sebe.
\newblock Pad-net: Multi-tasks guided prediction-and-distillation network for
  simultaneous depth estimation and scene parsing.
\newblock In {\em Proceedings of the IEEE Conference on Computer Vision and
  Pattern Recognition (CVPR)}, June 2018.

\bibitem{Yang_2018_ECCV}
Guorun Yang, Hengshuang Zhao, Jianping Shi, Zhidong Deng, and Jiaya Jia.
\newblock Segstereo: Exploiting semantic information for disparity estimation.
\newblock In {\em Proceedings of the European Conference on Computer Vision
  (ECCV)}, September 2018.

\bibitem{ocr_eccv_20}
Yuhui Yuan, Xilin Chen, and Jingdong Wang.
\newblock Object-contextual representations for semantic segmentation.
\newblock In Andrea Vedaldi, Horst Bischof, Thomas Brox, and Jan-Michael Frahm,
  editors, {\em Computer Vision -- ECCV 2020}, pages 173--190, Cham, 2020.
  Springer International Publishing.

\bibitem{Zhang_2018_ECCV}
Zhenyu Zhang, Zhen Cui, Chunyan Xu, Zequn Jie, Xiang Li, and Jian Yang.
\newblock Joint task-recursive learning for semantic segmentation and depth
  estimation.
\newblock In {\em Proceedings of the European Conference on Computer Vision
  (ECCV)}, September 2018.

\bibitem{Zhang_2019_CVPR}
Zhenyu Zhang, Zhen Cui, Chunyan Xu, Yan Yan, Nicu Sebe, and Jian Yang.
\newblock Pattern-affinitive propagation across depth, surface normal and
  semantic segmentation.
\newblock In {\em Proceedings of the IEEE/CVF Conference on Computer Vision and
  Pattern Recognition (CVPR)}, June 2019.

\bibitem{pspnet}
Hengshuang Zhao, Jianping Shi, Xiaojuan Qi, Xiaogang Wang, and Jiaya Jia.
\newblock Pyramid scene parsing network.
\newblock In {\em CVPR}, 2017.

\bibitem{domain_seg_nips_2}
Sicheng Zhao, Bo Li, Xiangyu Yue, Yang Gu, Pengfei Xu, Runbo Hu, Hua Chai, and
  Kurt Keutzer.
\newblock Multi-source domain adaptation for semantic segmentation.
\newblock In {\em Advances in Neural Information Processing Systems}, pages
  7285--7298, 2019.

\bibitem{zhao2020maskflownet}
Shengyu Zhao, Yilun Sheng, Yue Dong, Eric I-Chao Chang, and Yan Xu.
\newblock Maskflownet: Asymmetric feature matching with learnable occlusion
  mask.
\newblock In {\em Proceedings of the IEEE Conference on Computer Vision and
  Pattern Recognition (CVPR)}, 2020.

\bibitem{rmi_loss}
Shuai Zhao, Yang Wang, Zheng Yang, and Deng Cai.
\newblock Region mutual information loss for semantic segmentation.
\newblock In H. Wallach, H. Larochelle, A. Beygelzimer, F. d\textquotesingle
  Alch\'{e}-Buc, E. Fox, and R. Garnett, editors, {\em Advances in Neural
  Information Processing Systems}, volume~32, pages 11117--11127. Curran
  Associates, Inc., 2019.

\bibitem{lp_2006}
Xiaojin Zhu and Zoubin Ghahramani.
\newblock Learning from labeled and unlabeled data with label propagation.
\newblock In {\em CMU CALD tech report}, 2002.

\bibitem{nvidia_cvpr19}
Yi Zhu, Karan Sapra, Fitsum~A. Reda, Kevin~J. Shih, Shawn Newsam, Andrew Tao,
  and Bryan Catanzaro.
\newblock Improving semantic segmentation via video propagation and label
  relaxation.
\newblock In {\em The IEEE Conference on Computer Vision and Pattern
  Recognition (CVPR)}, June 2019.

\bibitem{domain_seg_8}
Yang Zou, Zhiding Yu, Xiaofeng Liu, B.V.K.~Vijaya Kumar, and Jinsong Wang.
\newblock Confidence regularized self-training.
\newblock In {\em The IEEE International Conference on Computer Vision (ICCV)},
  October 2019.

\bibitem{domain_seg_5}
Yang Zou, Zhiding Yu, B.V.K. Vijaya~Kumar, and Jinsong Wang.
\newblock Unsupervised domain adaptation for semantic segmentation via
  class-balanced self-training.
\newblock In {\em The European Conference on Computer Vision (ECCV)}, September
  2018.

\end{thebibliography}
